\documentclass{ieeeaccess}
\usepackage{cite}
\usepackage{amsmath,amssymb,amsfonts}
\usepackage{algorithmic}
\usepackage{graphicx}
\usepackage{textcomp}
\usepackage{booktabs} 
\usepackage{multirow} 
\usepackage{url}
\usepackage{enumitem}

\usepackage{bm}
\makeatletter
\AtBeginDocument{\DeclareMathVersion{bold}
\SetSymbolFont{operators}{bold}{T1}{times}{b}{n}
\SetSymbolFont{NewLetters}{bold}{T1}{times}{b}{it}
\SetMathAlphabet{\mathrm}{bold}{T1}{times}{b}{n}
\SetMathAlphabet{\mathit}{bold}{T1}{times}{b}{it}
\SetMathAlphabet{\mathbf}{bold}{T1}{times}{b}{n}
\SetMathAlphabet{\mathtt}{bold}{OT1}{pcr}{b}{n}
\SetSymbolFont{symbols}{bold}{OMS}{cmsy}{b}{n}
\renewcommand\boldmath{\@nomath\boldmath\mathversion{bold}}}
\makeatother

\def\BibTeX{{\rm B\kern-.05em{\sc i\kern-.025em b}\kern-.08em
    T\kern-.1667em\lower.7ex\hbox{E}\kern-.125emX}}

\DeclareFontShape{T1}{formata}{m}{sl}{<->ssub*formata/m/it}{}
\begin{document}
\history{This manuscript is currently under review at IEEE Access}
\doi{}

\title{Reflectance Multispectral Imaging for Soil Composition Estimation and USDA Texture Classification}
\author{\uppercase{G.A.S.L Ranasinghe}\authorrefmark{1}\textsuperscript{*},
\uppercase{J.A.S.T. Jayakody}\authorrefmark{2}\textsuperscript{*}, \uppercase{M.C.L. De Silva}\authorrefmark{1}, \uppercase{G. Thilakarathne}\authorrefmark{3}, \uppercase{G.M.R.I. GODALIYADDA}\authorrefmark{1,2}, \uppercase{H.M.V.R. HERATH}\authorrefmark{1,2}, \uppercase{M.P.B. EKANAYAKE}\authorrefmark{1,2}, and \uppercase{S.K. Navaratnarajah}\authorrefmark{4}}

\address[1]{Department of Electrical and Electronic Engineering, University of Peradeniya, Peradeniya 20400, Sri Lanka}
\address[2]{Multidisciplinary AI Research Center, University of Peradeniya, Peradeniya 20400, Sri Lanka}
\address[3]{Department of Electrical and Information Engineering, University of Ruhuna, Hapugala, Galle 80000, Sri Lanka}
\address[4]{Department of Civil Engineering, University of Peradeniya, Peradeniya 20400, Sri Lanka}
\tfootnote{\textsuperscript{*}G.A.S.L. Ranasinghe and J.A.S.T. Jayakody are co-first authors.}

\markboth
{S. Ranasinghe \headeretal: Reflectance Multispectral Imaging for Soil Texture Composition Estimation and USDA Texture Classification}
{S. Ranasinghe \headeretal: Reflectance Multispectral Imaging for Soil Texture Composition Estimation and USDA Texture Classification}

\corresp{Corresponding author: J.A.S.T. Jayakody (e-mail: senith@eng.pdn.ac.lk).}

\begin{abstract}
Soil texture is a foundational attribute that governs water availability and erosion in agriculture, as well as load‑bearing capacity, deformation response, and shrink–swell risk in geotechnical engineering. Yet texture is still typically determined by slow and labour intensive laboratory particle size tests, while many sensing alternatives are either costly or too coarse to support routine field scale deployment. This paper proposes a robust and field deployable multispectral imaging (MSI) system and machine learning framework for predicting soil composition and the United States Department of Agriculture (USDA) texture classes. The proposed system uses a cost effective in-house MSI device operating from 365 nm to 940 nm to capture thirteen spectral bands, which effectively capture the spectral properties of soil texture. Regression models use the captured spectral properties to estimate clay, silt, and sand percentages, while a direct classifier predicts one of the twelve USDA textural classes. Indirect classification is obtained by mapping the regressed compositions to texture classes via the USDA soil texture triangle. The framework is evaluated on mixture data by mixing clay, silt, and sand in varying proportions, using the USDA classification triangle as a basis. Experimental results show that the proposed approach achieves a coefficient of determination R\textsuperscript{2} up to 0.99 for composition prediction and over 99\% accuracy for texture classification. These findings indicate that MSI combined with data-driven modeling can provide accurate, non-destructive, and field deployable soil texture characterization suitable for geotechnical screening and precision agriculture.
\end{abstract}


\titlepgskip=-21pt

\maketitle

\section{Introduction}
\label{sec:introduction}

\PARstart{S}{oil} texture, which is the relative proportions of clay, silt, and sand, is a widely used descriptor that is routinely used to anticipate soil behavior across agricultural, environmental, and engineering systems \cite{Poggio2021SoilGrids2}. Texture influences how rapidly water infiltrates and drains and how much is retained for plants, making it a key input to hydrologic and land surface models \cite{Gupta2022GSHP,Wankmueller2024Nature}. Recent global analysis further indicates that texture related differences in water transmission shift the soil moisture conditions under which ecosystems become water limited, linking texture to drought sensitivity at broad scales \cite{Wankmueller2024Nature}. In infrastructure contexts, clay dominant soils can exhibit moisture induced volume change and strength variability, motivating conservative design or stabilization. The texture is also associated with mineral surface availability that stabilizes organic matter and influences soil carbon dynamics \cite{BarmanDash2022ExpansiveReview,Schweizer2021ClayMSA_SOC}. Collectively, these cross-domain roles make accurate soil texture determination essential for credible modeling and effective decision making in agriculture, environmental monitoring, hydrology, infrastructure design, and soil carbon assessment \cite{Gupta2022GSHP,Wankmueller2024Nature,BarmanDash2022ExpansiveReview,Schweizer2021ClayMSA_SOC}.

This importance is underscored by real world geotechnical failures and costly infrastructure damage when ground conditions are poorly characterized. The Highland Towers disaster in Kuala Lumpur (December 1993) illustrates how deficient site understanding can contribute to fatal slope failure during prolonged rainfall~\cite{Kazmi2017}. Likewise, shrink-swell cycles driven by moisture variations can damage foundations, pavements, and buried utilities, with reported annual losses of at least \$2.3~billion in the United States alone~\cite{HoltzHart1978SwellingSoils}. Similar reactive-soil issues remain a major concern in Australian practice, where residential slab and footing design explicitly accounts for expansive soil movement under seasonal wetting-drying cycles~\cite{Devkota2022,AS2870ReactiveSoils}. 

Due to the fundamental role of soil texture in agricultural management, scientific efforts to classify soils based on particle size distribution date back more than a century. One of the earliest structured classification proposals was introduced by Whitney in 1911~\cite{Whitney1911}, followed by refinements by Davis and Bennett in 1927~\cite{Davis1927}, ultimately leading to the standardized equilateral textural triangle adopted by the United States Department of Agriculture (USDA) in 1951~\cite{USDA1951} and later revised in subsequent editions of the Soil Survey Manual~\cite{USDA2017}. The USDA soil texture triangle remains the most widely used particle size based classification framework and defines twelve standard textural classes based on clay, silt, and sand percentages. Beyond its agricultural origins, the USDA texture system is extensively used in geotechnical engineering, hydrology, environmental science, and land resource management, where particle size distribution underpins assessments of permeability, compressibility, erosion susceptibility, and soil water dynamics~\cite{MorenoMaroto2022,Martin2018}.

Therefore, soil texture classification systems and the methods used to determine texture classes have continued to evolve from early mechanical analyses to modern data driven approaches~\cite{Whitney1911,Davis1927,USDA2017,Martin2018}. Traditionally, classification has been grounded in laboratory based particle size analysis, where the sand fraction is determined by sieve analysis and the silt and clay fractions are estimated using sedimentation techniques such as the hydrometer method based on Stokes’ law~\cite{Gee2018}.

To reduce reliance on laboratory particle size analysis for soil texture classification and characterization, sensing based approaches have been explored. At the low cost end, RGB imaging using digital or smartphone cameras has been investigated for soil texture classification using color/texture descriptors or learned features, but performance can be sensitive to illumination, surface roughness, and moisture driven appearance changes, limiting generalization outside controlled capture conditions~\cite{Barman2020SmartphoneSoilTexture,soil_classi_RGB_uncontr_field}. Proximal sensing methods, such as electromagnetic induction (EMI) and apparent electrical conductivity (ECa), are widely used for field scale mapping of texture related variability, particularly in clay rich soils. Yet, the measurements are indirect and strongly influenced by soil moisture and salinity, requiring careful interpretation and site specific calibration~\cite{EMI1,CorwinLesch2003AgronomyJ}. At broader spatial scales, spectroscopic approaches, including visible near-infrared (VNIR), mid-infrared (MIR), and hyperspectral imaging (HSI), have demonstrated strong capability for predicting clay content and particle size fractions when paired with chemometric or machine learning models. Still, routine deployment is often constrained by instrument cost, data volume, and strict acquisition requirements~\cite{7,8,issues}.

Recent advances in compact multispectral imaging (MSI) systems and data driven modeling provide a practical compromise between low cost RGB imaging and high dimensional hyperspectral sensing. Unlike RGB imaging, which records scene appearance using three broad visible channels, MSI acquires reflectance at multiple discrete wavelength bands spanning from near-ultraviolet (NUV) to near-infrared (NIR) and therefore provides richer spectral information for material discrimination~\cite{VanHoorn2024LowCostMSI}. By selecting a limited set of informative bands, MSI can reduce hardware and data burdens compared with hyperspectral sensing while retaining wavelength selective cues relevant to soil composition and texture related variation. In addition, MSI systems typically use controlled active illumination and a fixed optical setup, avoiding moving parts and bulky, fragile, or hypersensitive spectroscopic components. This design supports stable, repeatable measurements with minimal sample preparation, while improving durability and reducing cost. As a result, MSI can be implemented as a portable, field deployable platform that is often more robust in practice than HSI systems, remote sensing platforms, or laboratory spectroscopy based methods. Furthermore, MSI has demonstrated strong utility across agricultural and quality assessment applications, motivating its extension to soil characterization~\cite{19}.

The solution proposed in this paper leverages the inherent benefits of MSI, utilizing a customized thirteen band acquisition setup~\cite{21}, and employs a dedicated end to end learning pipeline to estimate soil clay, silt, and sand composition and the corresponding USDA texture class. Reflectance images are acquired across the selected spectral bands, and machine learning models are evaluated under a unified experimental protocol in three methodologies of characterizing soil texture: 

\begin{enumerate}
    \renewcommand{\labelenumi}{(\roman{enumi})}
    \item Direct classification of the twelve USDA texture classes from multispectral features 
    \item Regression to estimate clay, silt, and sand percentages from the same features
    \item Indirect classification, where the predicted composition from case (ii) is mapped to USDA texture classes using standard USDA texture triangle decision rules. 
\end{enumerate} 

By benchmarking these three methods, the work quantifies the accuracy and practical trade offs of MSI based soil texture assessment and supports the development of compact, field deployable tools for geotechnical, agricultural, environmental, or any other pertinent decision making application. 

The main contributions of the work are as follows. 

\begin{itemize} 
    \item A cost effective, field deployable thirteen band MSI acquisition and preprocessing workflow for soil reflectance characterization.
    \item A machine learning based pipeline for estimating soil composition in terms of clay, silt, and sand fractions.
    \item A dual soil texture classification framework: (a) direct classification of USDA texture classes, (b) indirect classification via regression based soil composition mapped onto the texture triangle.
    \item A curated soil reflectance dataset constructed from laboratory prepared clay, silt, and sand mixtures with ground truth particle size composition and corresponding USDA texture labels.
\end{itemize}

The remainder of this paper is organized as follows, Section~\ref{sec:related_work} reviews imaging and sensing based approaches for soil texture estimation. Section~\ref{sec:MatMethods} describes the proposed MSI acquisition system and modeling framework. Section~\ref{sec:results} presents the results for direct classification, composition regression, and indirect triangle based classification. Section~\ref{sec:discussion} discusses the findings and trade offs between direct and indirect pipelines. Finally, Section~\ref{sec:conclusion} concludes the paper.

\section{Related Work}
\label{sec:related_work}

RGB and conventional imaging techniques have been investigated as low cost approaches for soil texture classification. However, most are limited in the number of texture classes and do not support continuous estimation of individual soil texture constituents (clay, silt, and sand). The reliable quantification of these constituent fractions is critical for comprehensive granulometric characterization and process based soil modeling \cite{granbook}. One study captured smartphone photographs of soil samples under controlled lighting and extracted color moments, Gabor wavelet features, and HSV histograms for direct classification of USDA soil texture classes using a multiclass support vector machine (SVM) classifier~\cite{Barman2020SmartphoneSoilTexture}. Another work employed multiple machine learning methods to classify seven texture based soil classes from RGB images~\cite{spider_monkey}, while a further study directly classified soil samples into three coarse USDA based soil classes using RGB images and deep learning~\cite{soil_classi_RGB_uncontr_field}. These RGB based methods are rapid and inexpensive, but highly dependent on lighting conditions and do not estimate continuous clay, silt, and sand composition. 

Laboratory based HSI has been widely used for soil composition prediction, typically leveraging large spectral libraries. Most studies rely on the LUCAS (a European dataset), Australian, Brazilian, and Kenyan soil datasets, with reflectance spectra typically ranging from 350 nm to 2800 nm at resolutions ranging from 0.5 nm to 2 nm \cite{deepL_regional_spectral, joint_calibration_geoderma, soil_land_assesment_geoderma, simultaneous_multi_property_geoderma, on-site_analysis_geoderma, laser_spectro_geoderma}. These studies demonstrate that laboratory HSI can accurately predict soil texture fractions and related properties. However, most of them do not explicitly classify soil texture classes. Furthermore, the majority of these works depend on widely available datasets rather than developing dedicated HSI image data from newly acquired soil samples. This reliance may stem from the high cost and limited accessibility associated with laboratory-grade HSI systems.

Satellite based remote sensing has been widely used for regional-scale soil texture mapping using multispectral and, in some cases, hyperspectral imagery. Studies commonly utilize data from Sentinel-2, Sentinel-3, Landsat-8 OLI, and airborne hyperspectral sensors, covering spectral ranges approximately between 370 nm and 2440 nm, to predict clay, silt, and sand fractions or to classify texture classes \cite{soil_mapping_air_sentinel_2, remote_sensing_texture_classes, soil_charac_sentinel_3, predict_soil_texture_Jianghuai, MSI_remote_particle_size_Wuhai}. These approaches often integrate spectral indices such as NDVI and auxiliary information including topographic and geological data to improve prediction performance. While satellite imaging enables large-scale mapping, it is limited by coarse spatial resolution and by the requirement for bare soil pixels, which constrains its applicability in many agricultural and natural settings.

Portable MSI systems have been explored far less frequently for soil texture estimation, despite their potential for low cost, field deployable sensing. Based on current literature, only one study has directly classified mixtures of clay, silt, and sand using a seven band MSI system~\cite{soil_texture_class_USDA}. This work demonstrates that compact MSI can distinguish predefined mixtures, but it neither estimates soil composition nor indirectly classifies soil types according to the USDA soil texture triangle. These limitations highlight a gap for portable MSI systems that can support both soil composition prediction and USDA texture classification.

\section{Material and Methods}
\label{sec:MatMethods}

\subsection{Multispectral Imaging System}
\label{sub:MSI_Sys}
Although commercial MSI systems offer high spectral and spatial resolution, they are often prohibitively expensive for many laboratory and field deployments. Moreover, acquiring dense spectral measurements may be unnecessary when the target material exhibits a limited discriminative response in certain spectral regions, thereby introducing redundancy and increasing processing costs. To address these constraints, a custom low cost MSI system was developed in-house~\cite{21}. The system captures images at a selected set of wavelengths from NUV to NIR, balancing affordability with task specific spectral sensitivity. This MSI platform has already been successfully utilized for soil moisture estimation~\cite{19}, demonstrating its applicability for soil texture estimation in this work.

\begin{figure}[tb]
    \centering
    \includegraphics[width=0.9\linewidth]{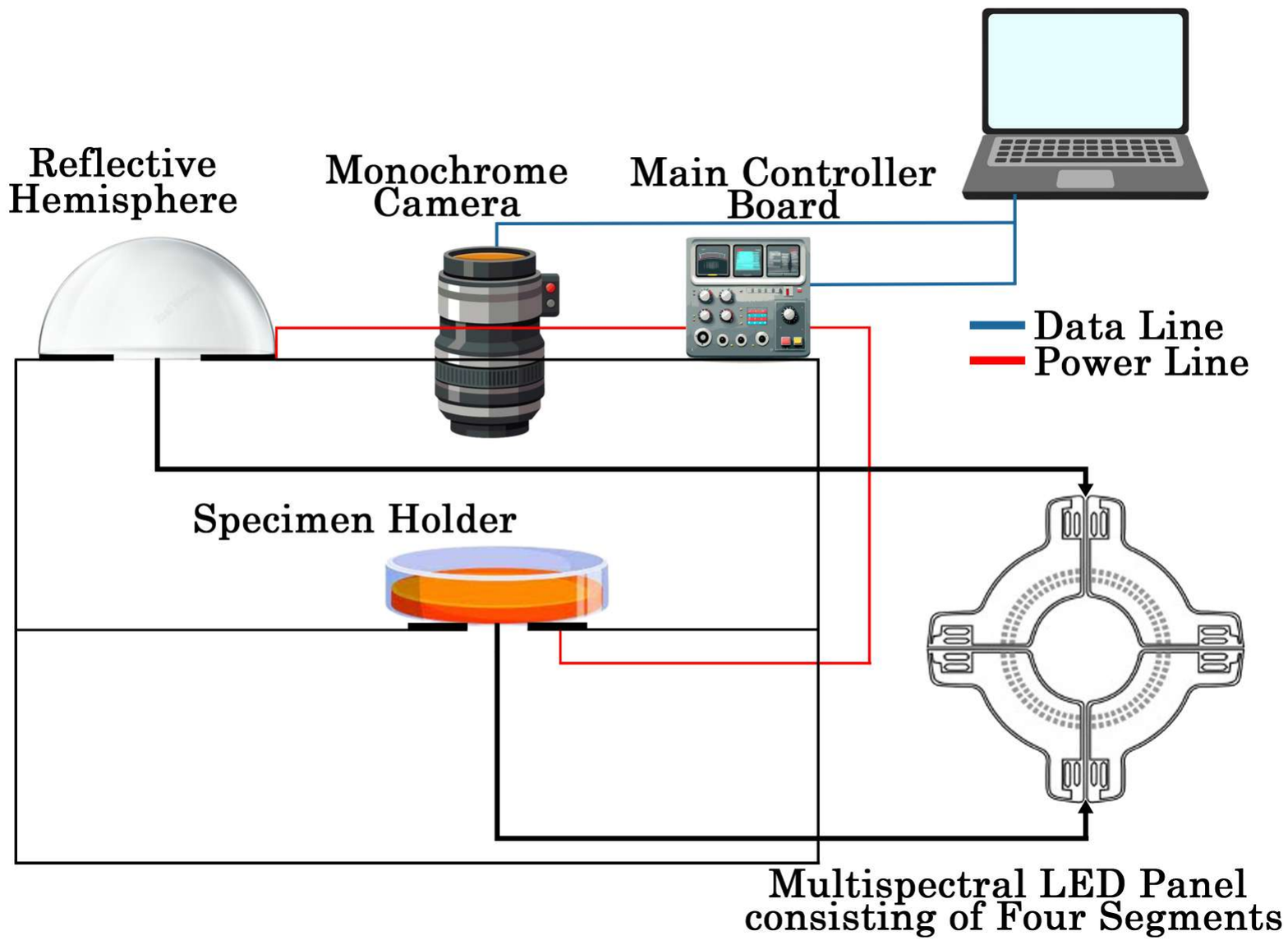}
    \caption{Schematic of the multispectral imaging system.}
    \label{msi_device}
\end{figure}

As shown in Fig.~\ref{msi_device}, image acquisition is performed using a FLIR BFS-U3-13Y3M CMOS monochrome machine vision camera (1.3~MP, USB3 Vision v1.0, 1280$\times$1024 resolution, 10-bit ADC) equipped with an onsemi PYTHON1300 image sensor. Illumination is provided at thirteen central wavelengths of 365, 405, 473, 530, 575, 621, 660, 735, 770, 830, 850, 890, and 940 nm, with each wavelength band generated by a dedicated cluster of narrowband LEDs. These bands were selected to provide broad spectral coverage while maintaining hardware simplicity and cost efficiency.

To ensure repeatable measurements, the camera, illumination panel, and sample holder are enclosed in a custom built dark chamber to minimize interference from ambient light and external reflections. The interior surfaces are coated with a nonreflective matte black material to suppress stray light, supporting consistent imaging across spectral bands. The effective spectral operating range of the system (350-1080~nm) aligns with the camera sensitivity profile. LED switching and acquisition timing are controlled using an Arduino Due microcontroller, while a connected host computer handles image transfer and processing.

\subsection{Soil Sample Preparation}
\label{subs:Soil_Samples}
\begin{figure}[tb]
    \centering
    \includegraphics[width=0.80\linewidth]{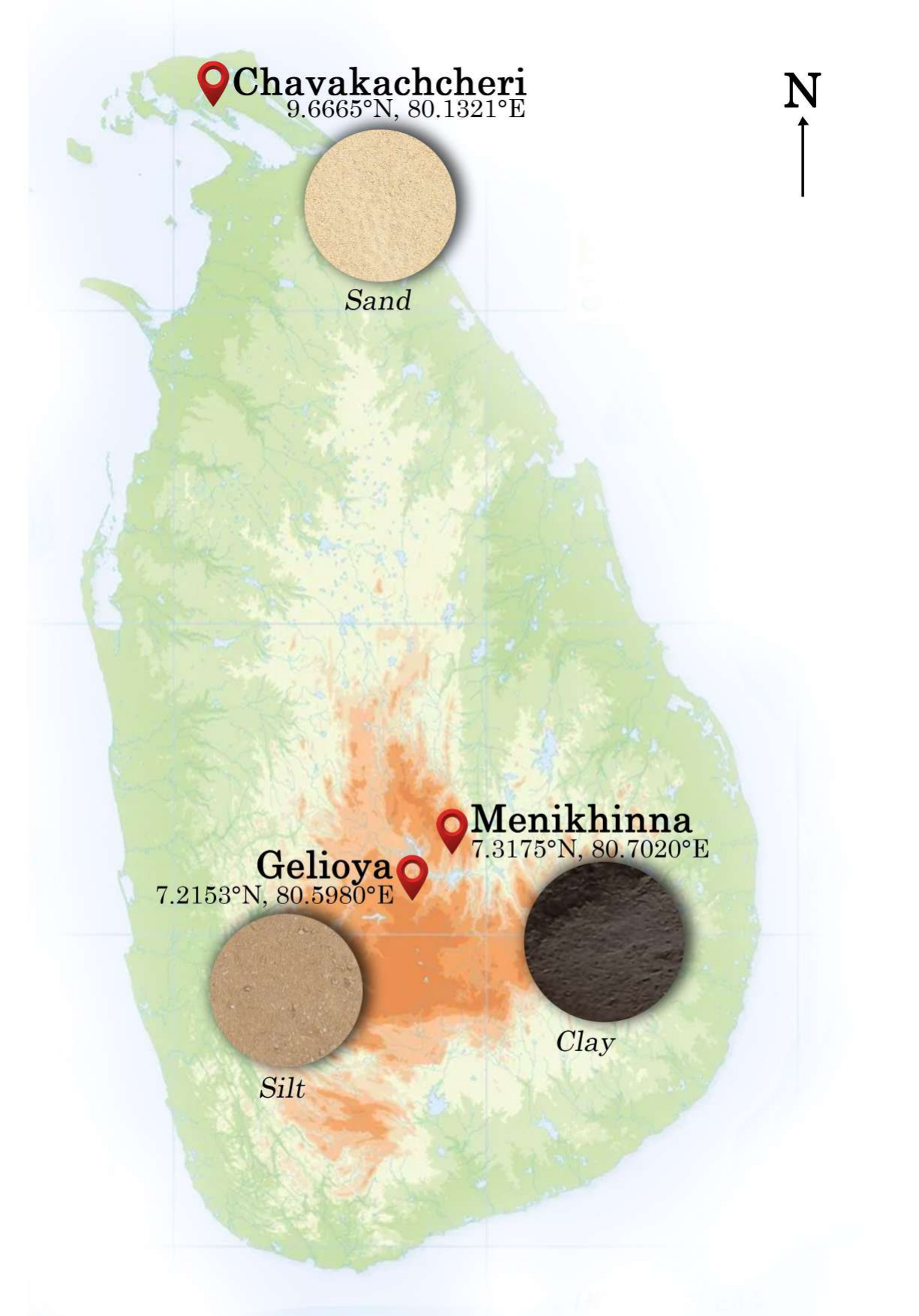}
    \caption{Map of Sri Lanka showing the locations of soil collection sites: clay rich soil from Menikhinna, silt rich soil from Gelioya, and sand rich soil from Chavakachcheri. Coordinates are provided in WGS84.}
    \label{fig:soil_map}
\end{figure}

To evaluate the proposed approach, soil samples with known particle size composition and corresponding USDA soil texture classes were required. Because naturally occurring soils that perfectly represent pure clay, silt, and sand are difficult to locate, three representative source soils with dominant clay, silt, and sand fractions were selected as near endmembers. Clay rich soil was collected from Menikhinna, silt rich soil from Gelioya, and sand rich soil from Chavakachcheri, located in different regions of Sri Lanka. The collection sites were chosen to provide representative endmembers across distinct physiographic regions of Sri Lanka and are shown in Fig.~\ref{fig:soil_map}.

After collection, soils were manually cleaned to remove gravel, stones, roots, and other organic debris. To standardize moisture conditions before processing, all soils were oven dried at 105\,\textdegree C for 24~h~\cite{18}. Each sample was then sieved through a 2~mm mesh to isolate the fine earth fraction, following standard soil preparation procedures~\cite{USDA2017,sieve}. The particle size composition of the source soils was confirmed using sieve and hydrometer analyses~\cite{4,17}, as detailed in Section~\ref{subs:lab_test}.

\begin{figure}[tb]
    \centering
    \includegraphics[width=1.0\linewidth]{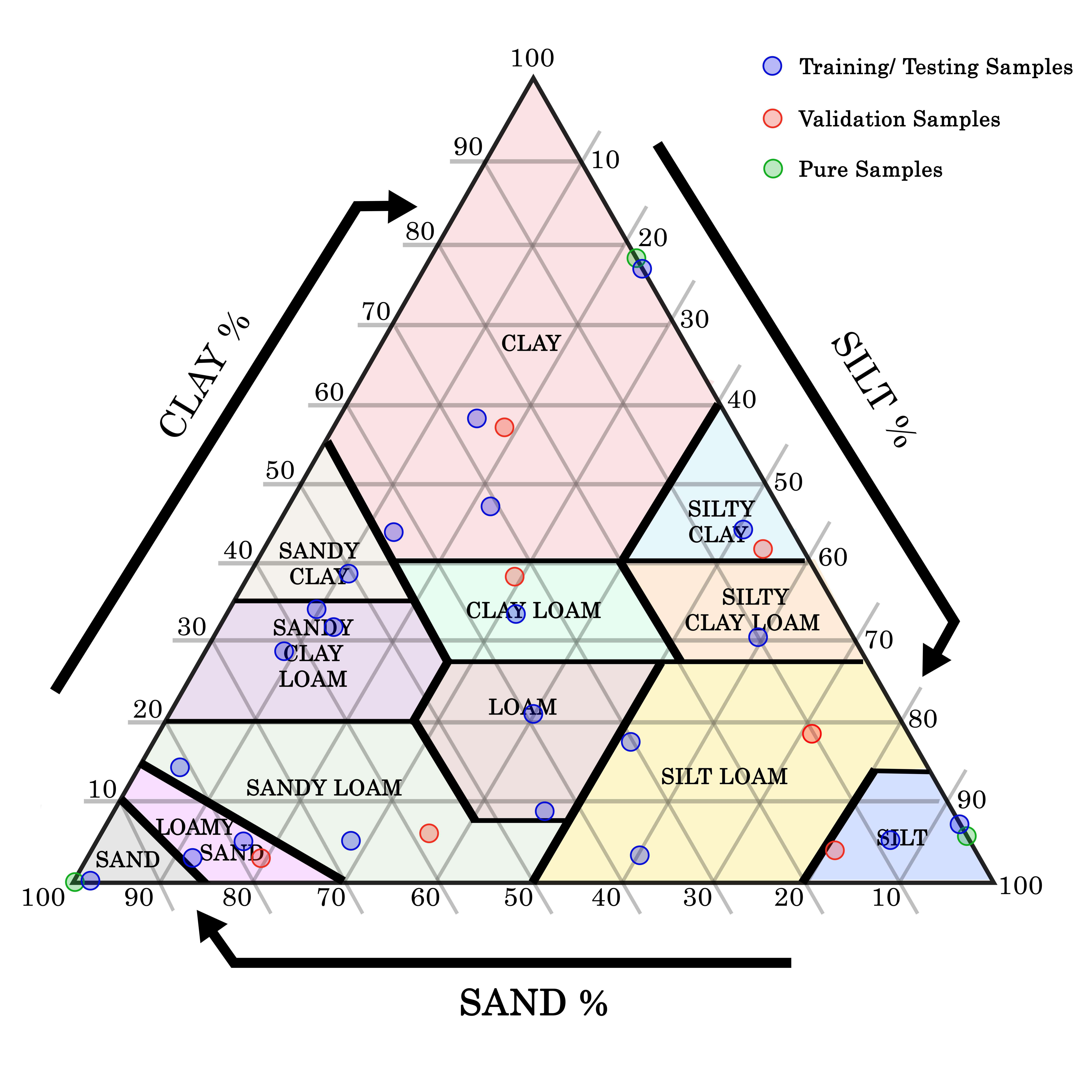}
    \caption{USDA soil texture triangle showing the distribution of samples used in this study. Green dots indicate the three source soils (clay, silt, and sand rich endmembers), blue dots indicate mixtures used for model development (training/testing), and red dots indicate mixtures used for external validation.}
    \label{fig:texture_triangle}
\end{figure}

Composite specimens were then prepared by mixing the three source soils in controlled mass ratios to cover all twelve USDA soil texture classes. The distribution of the prepared mixtures within the USDA soil texture triangle is illustrated in Fig.~\ref{fig:texture_triangle}. Each mixture was thoroughly homogenized and placed in aluminum containers (5 cm in diameter, 4 cm in height). For the dataset (training/testing), 22 unique mixture ratios were prepared, with 20 replicates per ratio, yielding 440 soil specimens. For external validation, 7 additional mixture ratios were prepared using intermediate points in the USDA triangle as reference, with 12 replicates per ratio, yielding 84 specimens. In total, 524 specimens were prepared. 

\subsection{Laboratory Test for Composition}
\label{subs:lab_test}
To obtain ground truth particle size composition for the collected clay rich, silt rich, and sand rich source soils, laboratory particle size analysis was conducted. Mechanical analysis was performed using sieve analysis and hydrometer analysis in combination~\cite{4, iso112772009}. Sieve analysis characterizes the coarse fraction (particles larger than 0.075\,mm), whereas hydrometer analysis is applied to the fine fraction (particles smaller than 0.075\,mm)~\cite{Reynolds2002}. Following the ASTM D2487~\cite{ASTM_D2487_2017} standard, particles smaller than 0.002\,mm were classified as clay, particles between 0.002\,mm and 0.075\,mm as silt, and particles between 0.075\,mm and 2\,mm as sand~\cite{USDA2017}. Particles larger than 2\,mm were excluded from the soil fraction as described in Section~\ref{subs:Soil_Samples}. The final percentages of clay, silt, and sand were obtained from the combined particle size distribution derived from sieve and hydrometer measurements.

For each soil type, an 80\,g air dried specimen was used for sieve analysis. A stack of sieves with progressively smaller apertures was employed. The mass retained on each sieve was recorded and used to compute cumulative passing percentages. Since gravel and other coarse fragments had been removed during sample preparation, this procedure provided the sand fraction of each soil~\cite{USDA2017}.

The fine fraction passing the 0.075\,mm sieve was subsequently analyzed using the hydrometer method following Bouyoucos~\cite{bouyoucos1962hydrometer}. This technique estimates the particle size distribution of soil suspended in a fluid based on Stokes' law, whereby smaller sized particles settle more slowly than larger sized particles. Sodium hexametaphosphate ($(NaPO_3)_6$) and sodium carbonate ($Na_2CO_3$) were used to disperse the fine soil particles by increasing the suspension pH and promoting electrostatic repulsion of fine particles~\cite{dispersing_agents}, and the suspension volume was adjusted to 1000\,ml with distilled water in a sedimentation cylinder. Hydrometer and temperature readings were taken at specified intervals over a 24~h period, starting from the initial (zero) reading. Finally, the suspension was dried and weighed to verify mass recovery. Using these measurements, the proportions of silt and clay were estimated. Overall, the complete mechanical analysis required approximately 1-2 days.

Table~\ref{tab:endmember_composition} reports the measured particle size composition of the three source soils. Using these laboratory-measured percentages of clay, silt, and sand in source soils, ground truth targets for all soil mixture ratios were derived for the regression experiments. The corresponding USDA soil texture labels were assigned by mapping each mixture’s clay, silt, and sand percentages onto the USDA texture triangle using its standard decision boundaries, and these labels were used as ground truth for the classification experiments.

\begin{table}[tb]
\centering
\caption{Laboratory-measured particle size composition of the source soils.}
\label{tab:endmember_composition}
\small
\begin{tabular}{lccc}
\hline
\textbf{Source soil} & \textbf{Clay (\%)} & \textbf{Silt (\%)} & \textbf{Sand (\%)}\\
\hline
Clay rich soil & 78.63 & 21.37 & 0.00\\
Sand rich soil & 0.00  & 0.00  & 100.00\\
Silt rich soil & 5.75  & 94.25 & 0.00\\
\hline
\end{tabular}
\end{table}

\subsection{Image Acquisition}
\label{subs:Img_Acq}
Before data collection, the imaging system was configured to ensure consistent acquisition across all samples. The camera focus was adjusted to resolve surface texture clearly, and the aperture and exposure settings were tuned to avoid saturation while maintaining adequate signal across all wavelength bands. The field of view was set to fully cover the soil surface within the container while minimizing the area of background. After this initial setup, the imaging geometry and camera settings were kept fixed throughout the experiment to ensure repeatability.

\begin{figure}[tb]
    \centering
    \includegraphics[width=1.0\linewidth]{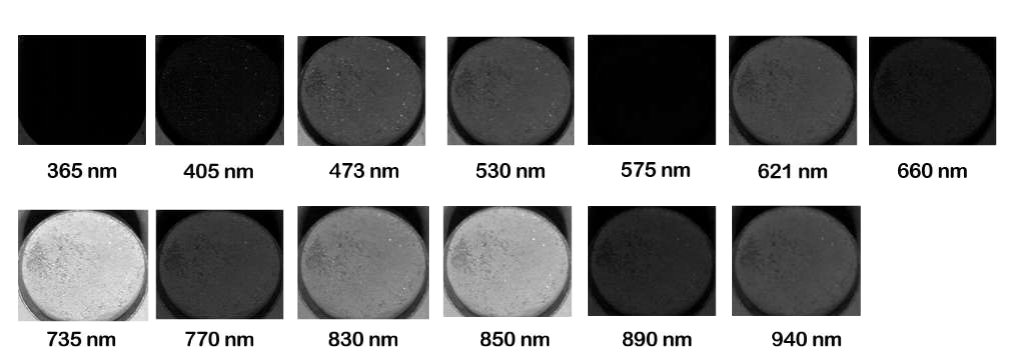}
    \caption{Representative monochrome spectral band images of a soil sample acquired using the proposed MSI system under thirteen narrow band LED illuminations (one image per wavelength).}
    \label{spectal_img}
\end{figure}

As described in Section~\ref{subs:Soil_Samples}, each soil mixture was placed in an aluminum container. For imaging, the container was fixed in place within the dark chamber using a consistent placement guide, and the sample surface was leveled to minimize shading. For each specimen, a multispectral image cube was acquired by sequentially illuminating the sample with the thirteen narrowband LEDs and capturing one monochrome image at each wavelength band. The same acquisition protocol was applied to all samples and replicates used in this study, including those reserved for external validation. Fig.~\ref{spectal_img} illustrates representative monochrome images of a single soil sample captured at the different wavelength bands.

\subsection{Image Preprocessing}
\label{subs:img_pre}
\begin{figure}[tb]
    \centering
    \includegraphics[width=1.0\linewidth]{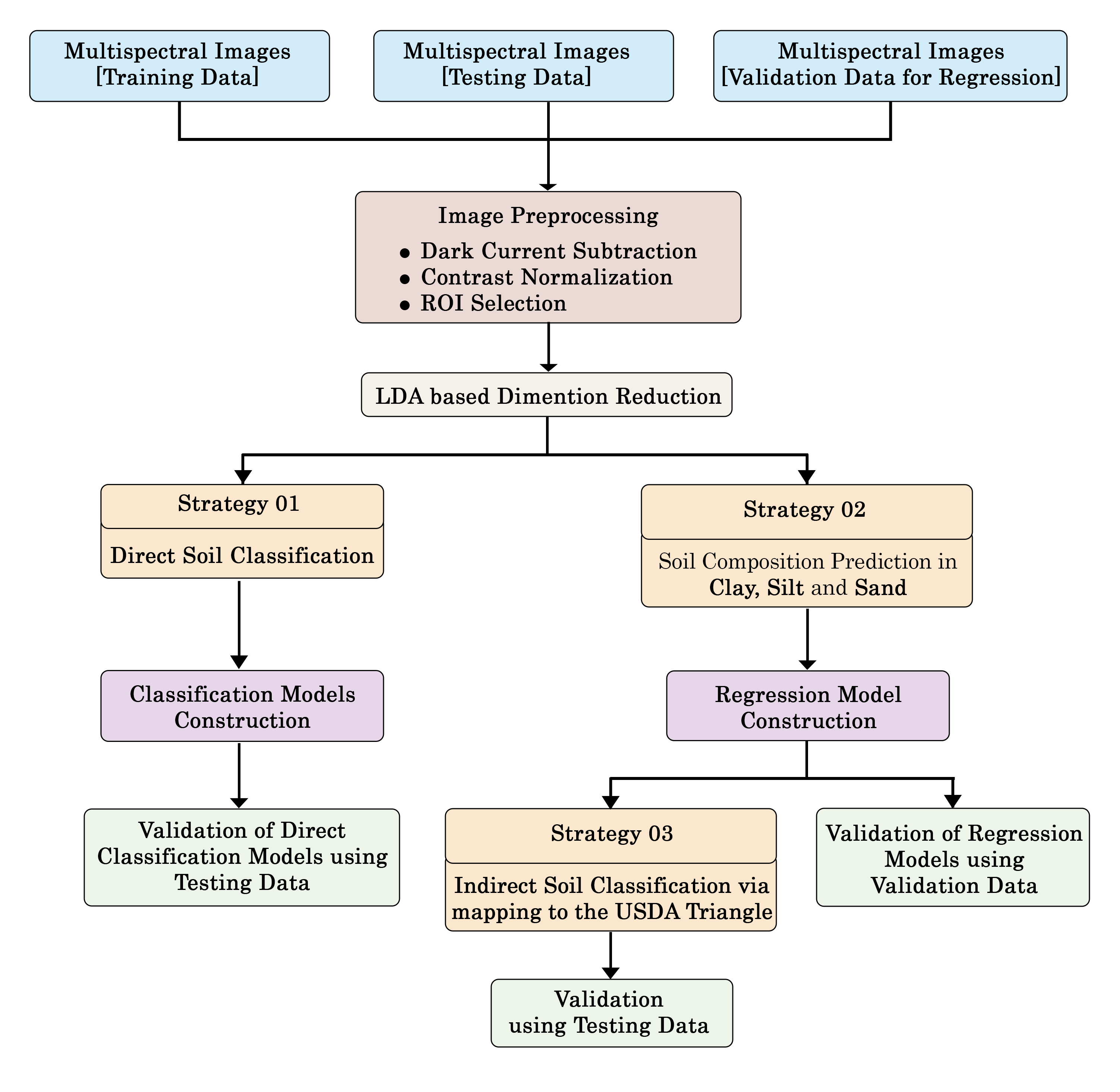}
    \caption{Workflow of the proposed MSI based pipeline. After preprocessing and LDA based dimensionality reduction, three studies are evaluated: (Strategy~1) direct USDA texture class classification, (Strategy~2) regression based clay, silt, and sand prediction, and (Strategy~3) indirect USDA texture classification via texture triangle mapping, using the designated test/validation splits.}
    \label{fig:workflow}
\end{figure}

Following image acquisition (Section~\ref{subs:Img_Acq}), each multispectral sample was processed using a standardized pipeline before feature extraction and modeling. The overall workflow, including the two downstream modeling strategies evaluated in this study, is summarized in Fig.~\ref{fig:workflow}. In brief, preprocessing comprised (1) dark current correction, (2) region of interest (ROI) selection and cropping, and (3) contrast normalization applied to the dark corrected, cropped ROI.

\subsubsection{Dark current correction}
\label{subsub:dark_correction}
Imaging sensors exhibit dark current and fixed pattern noise even in the absence of illumination. To reduce these effects, a dark frame $D$ was acquired with the illumination off and used to correct each spectral band image $X(\lambda)$. In this work, dark correction was implemented using an absolute difference operation:
\begin{equation}
Y(\lambda) = \left|\, X(\lambda) - D \,\right|,
\label{eq:dark_absdiff}
\end{equation}
where $X(\lambda)$ denotes the captured image at wavelength $\lambda$ and $Y(\lambda)$ denotes the dark corrected image. The absolute operator ensures non negative intensities and prevents negative pixel values that may arise from sensor noise and digitization offsets.

\subsubsection{ROI selection and cropping}
\label{subsub:roi_crop}
A fixed ROI $\Omega$, was selected such that the soil sample was fully contained while minimizing background pixels. To preserve spectral alignment, the same ROI coordinates were applied across all wavelength bands. Let $(x_1, y_1)$ denote the top left corner of the ROI. Cropping was performed as
\begin{equation}
Y_{\Omega}(\lambda) = Y(\lambda)\big[y_1 : y_1 + 100,\; x_1 : x_1 + 100\big],
\label{eq:crop}
\end{equation}
resulting in a fixed $100 \times 100$ pixel ROI for every spectral band and sample.

\subsubsection{Contrast normalization on the dark corrected ROI}
\label{subsub:hist_eq}
To enhance contrast while limiting the influence of outliers, a bounded non linear intensity mapping was applied independently to each dark corrected cropped ROI $Y_{\Omega}(\lambda)$. Similar bounded, sigmoidal mappings have been widely used in image preprocessing to achieve controlled contrast normalization while suppressing extreme intensity values~\cite{PIZER1987355,109340, PerezEnriquez2023Sigmoid}.  

For each wavelength band, the mean and standard deviation within the ROI were computed as
\begin{equation}
\mu_\lambda = \frac{1}{N}\sum_{p \in \Omega} Y_{\Omega}(\lambda,p),
\qquad
\sigma_\lambda = \sqrt{\frac{1}{N}\sum_{p \in \Omega}\left(Y_{\Omega}(\lambda,p)-\mu_\lambda\right)^2},
\label{eq:mean_std}
\end{equation}
where $\Omega$ contains $N$ pixels while $p$ indicates pixel locations.

A smooth, bounded mapping based on the hyperbolic tangent function was then applied as
\begin{equation}
\hat{Y}_{\Omega}(\lambda,p) =
\left(\mu_\lambda-\sigma_\lambda\right)
+ 2\sigma_\lambda \cdot 
\frac{\tanh\!\left(\kappa\left(Y_{\Omega}(\lambda,p)-\mu_\lambda\right)\right)+1}{2},
\label{eq:tanh_eq}
\end{equation}
where $\kappa$ controls the steepness of the transformation (set to $\kappa=0.03$). Hyperbolic tangent based mappings provide smooth compression of extreme values while preserving mid range intensity variation, making them well suited for robust normalization of reflectance data~\cite{Ouyang2010AIHT}.

\begin{figure}[tb]
    \centering
    \includegraphics[width=1.0\linewidth]{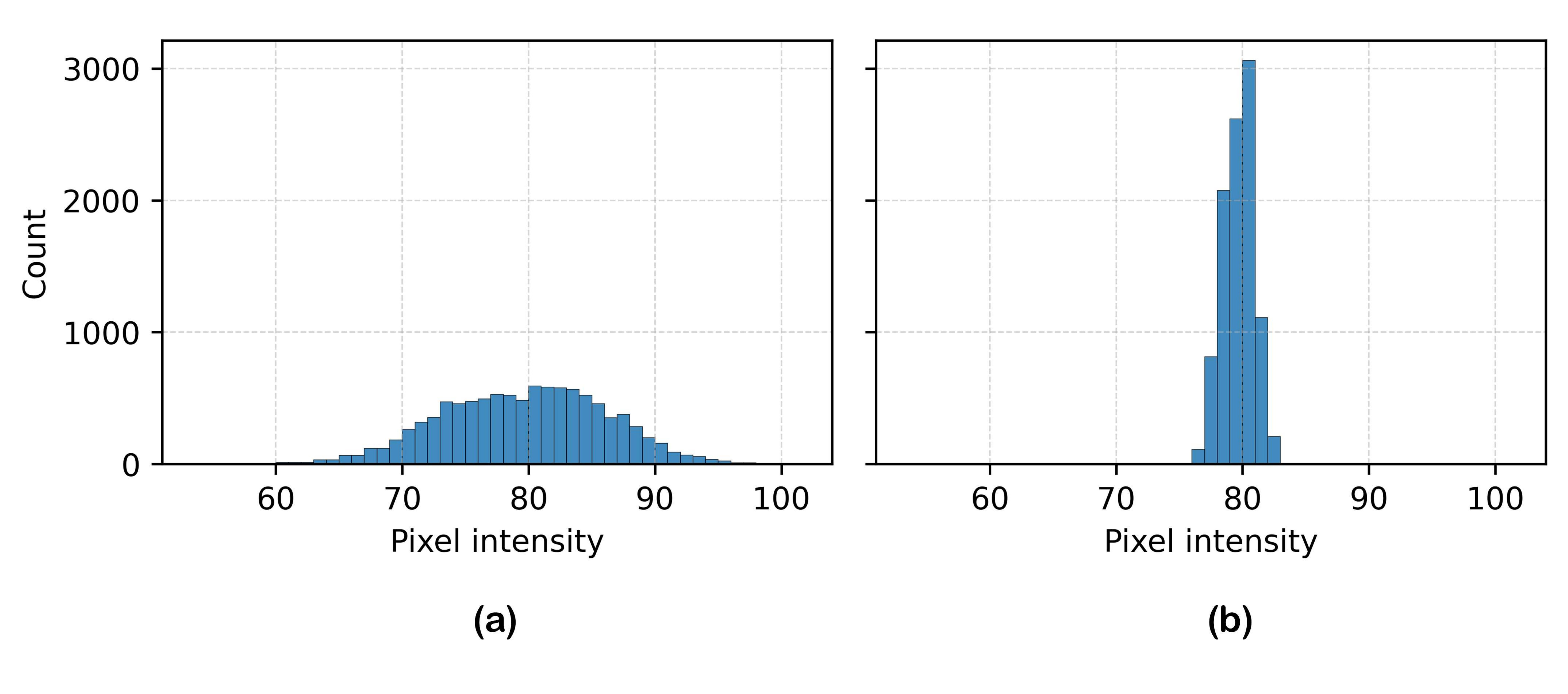}
    \caption{Representative ROI intensity histograms at $\lambda = 735$\,nm, ~~~~~~~~~~(a) before and (b) after the proposed bounded contrast normalization, computed from the dark corrected ROI. The same preprocessing is applied to all bands and samples.}
    \label{fig:hist_735_single}
\end{figure}

Since $\tanh(\cdot)\in[-1,1]$, the mapping constrains intensities approximately to
\begin{equation}
\hat{Y}_{\Omega}(\lambda,p) \in \left[\mu_\lambda-\sigma_\lambda,\; \mu_\lambda+\sigma_\lambda\right],
\label{eq:range}
\end{equation}
thereby providing controlled normalization and consistent scaling within each spectral band. Fig.~\ref{fig:hist_735_single} illustrates the effect of the proposed normalization using a representative example at $\lambda = 735$\,nm. Before normalization (Fig.~\ref{fig:hist_735_single} (a)), the pixel intensities exhibit a broader distribution with noticeable spread and mild tail behavior, reflecting intra ROI variability and potential influence of extreme values. After applying the bounded hyperbolic tangent mapping (Fig.~\ref{fig:hist_735_single} (b)), the distribution becomes more compact and concentrated around the mean, with the dynamic range effectively constrained within approximately $\mu_\lambda \pm \sigma_\lambda$. Extreme intensity values are smoothly compressed rather than abruptly clipped, while mid range variations remain distinguishable. This controlled redistribution enhances contrast consistency across samples and bands, reduces sensitivity to outliers, and promotes stable downstream feature extraction.

\subsection{Spectral Feature Extraction}
\label{subs:feat_ext}
After preprocessing (Section~\ref{subs:img_pre}), spectral features were extracted from each multispectral cube comprising thirteen wavelength bands. Let $\hat{Y}_{\Omega}(\lambda_i)$ denote the preprocessed $100{\times}100$ pixel ROI corresponding to wavelength $\lambda_i$.

\begin{figure}[tb]
    \centering
    \includegraphics[width=1.0\linewidth]{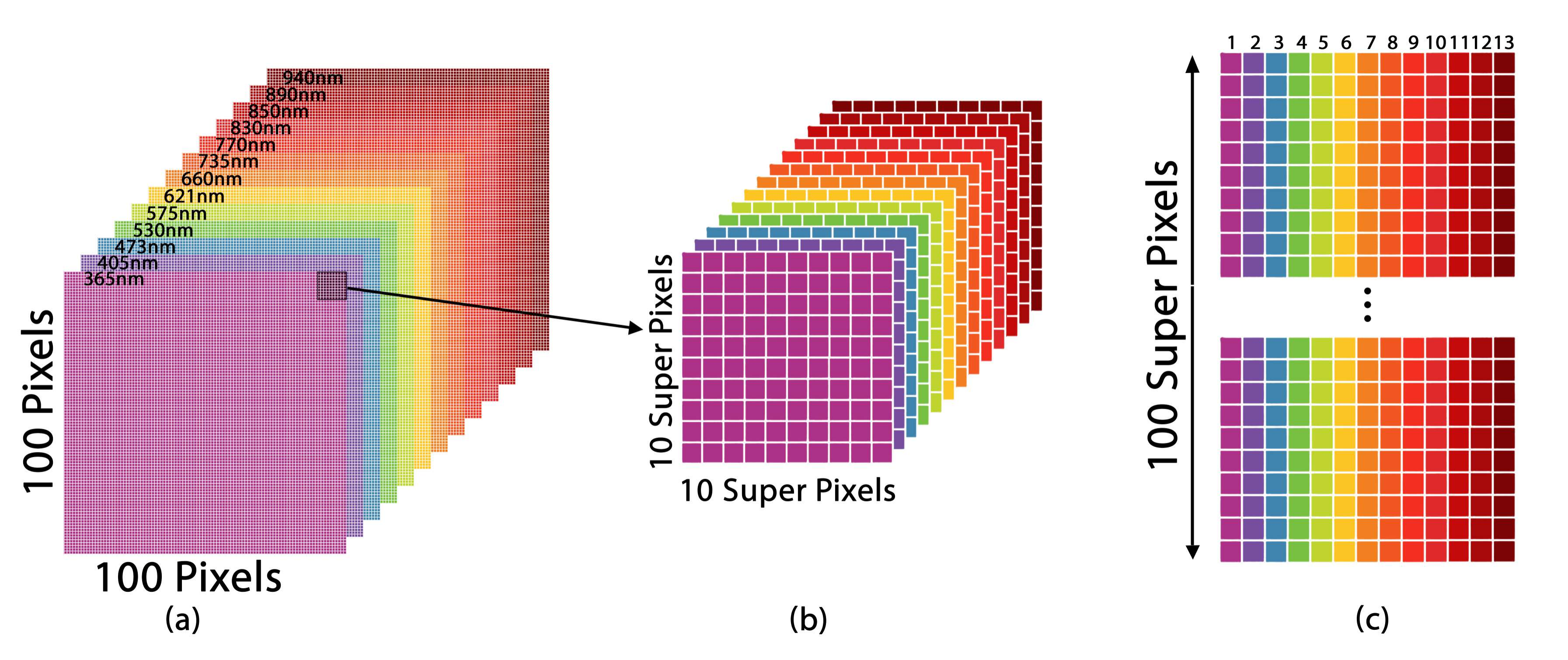}
    \caption{Spectral feature extraction from the preprocessed $100{\times}100$ pixel ROI. The ROI is partitioned into $10{\times}10$ blocks, and block wise mean intensities are computed for each of the thirteen wavelength bands to form a per sample feature matrix $\mathbf{X}\in\mathbb{R}^{100\times13}$.}
    \label{fig:feat_extraction}
\end{figure}

To obtain a robust representation while capturing intra sample variability, the ROI was partitioned into a regular grid of $10{\times}10$ non-overlapping blocks, each of size $10{\times}10$ pixels (Fig.~\ref{fig:feat_extraction}). This block based feature extraction was applied consistently to the dataset (training/testing) as well as the external validation dataset. For each wavelength band $\lambda_i$, the mean intensity of block $(u,v)$ was computed as
\begin{equation}
\bar{R}_{i}(u,v)=\frac{1}{|\mathcal{B}_{u,v}|}\sum_{p\in\mathcal{B}_{u,v}}\hat{Y}_{\Omega}(\lambda_i,p),
\label{eq:block_mean}
\end{equation}
where $u,v\in\{1,\ldots,10\}$ index the block row and column, $\mathcal{B}_{u,v}$ denotes the set of pixels in block $(u,v)$, and $|\mathcal{B}_{u,v}|=100$. The block means for each band were vectorized to form a hundred-dimensional vector,
\begin{equation}
\mathbf{x}_{i}=\left[\,\bar{R}_{i}(1,1),\ldots,\bar{R}_{i}(10,10)\,\right]^{\mathsf{T}},
\label{eq:xi}
\end{equation}
and the per sample multispectral feature matrix was constructed by concatenating the vectors from all thirteen bands,
\begin{equation}
\mathbf{X}=\left[\,\mathbf{x}_{1},\mathbf{x}_{2},\ldots,\mathbf{x}_{13}\,\right]\in\mathbb{R}^{100\times 13}.
\label{eq:Xmat}
\end{equation}

The final learning dataset was formed by vertically concatenating the block level observations from all samples, yielding a matrix with thirteen spectral features per observation. Since each specimen yields $10\times10=100$ non-overlapping blocks, the dataset (training/testing) corresponds to 44\,000 block level observations (440 specimens $\times$ 100 blocks), and the external validation set corresponds to 8400 block level observations (84 specimens $\times$ 100 blocks).

\textit{Feature normalization for modeling:}
Prior to dimensionality reduction and machine learning, each spectral feature (each wavelength band) was scaled using min–max normalization based on the training/testing dataset. Specifically, for each band, feature values were linearly mapped to the $[0,1]$ range using the minimum and maximum values. The same scaling parameters were then applied to the corresponding external validation dataset.

\subsection{Spectral Signature Visualization}
\label{subs:spectral_sig}
\begin{figure*}[tb]
    \centering
    \includegraphics[width=\textwidth]{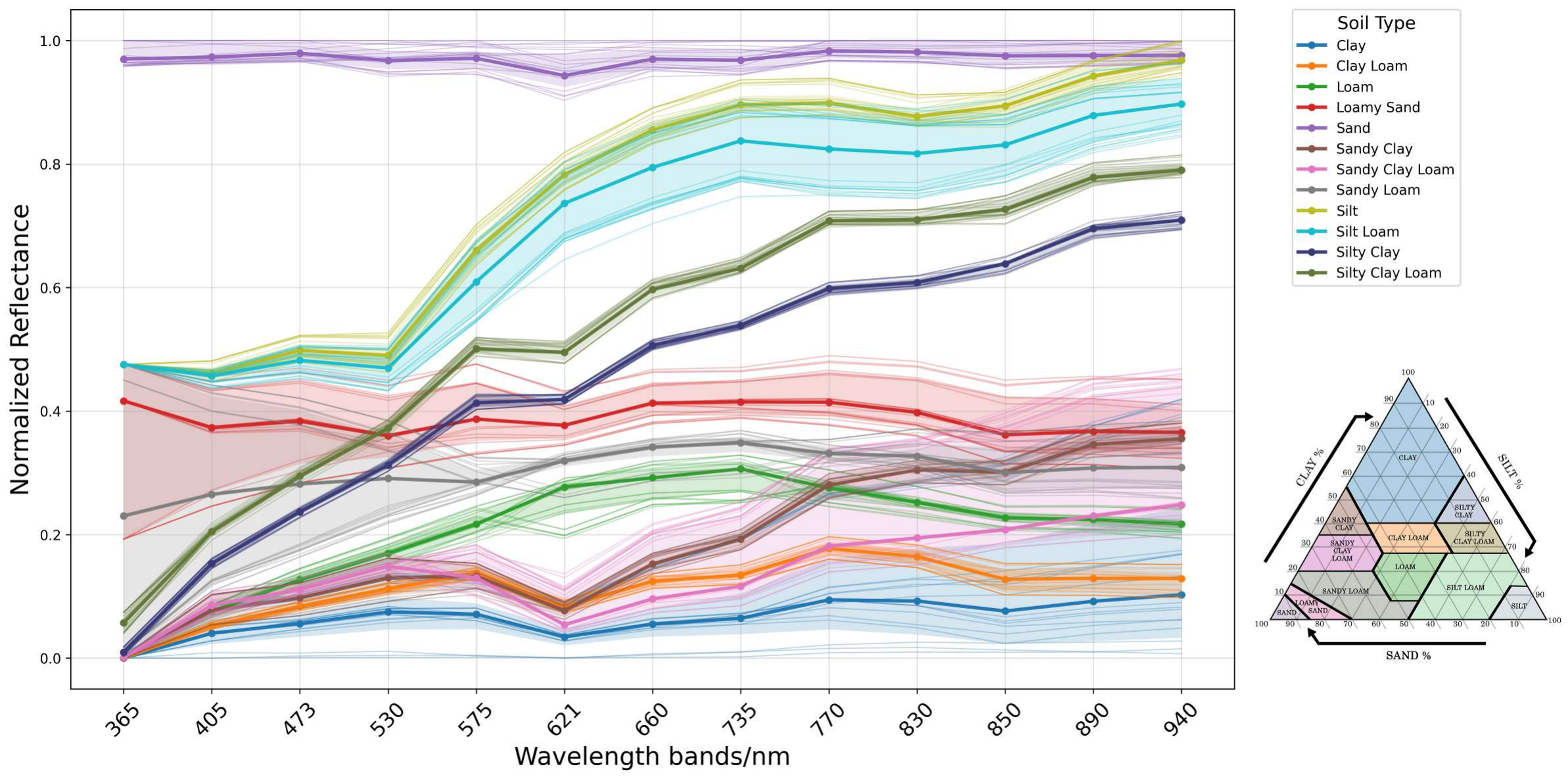}
    \caption{Spectral signatures grouped by the twelve USDA soil texture classes using the training/testing dataset. Each curve represents the mean intensity of the $100{\times}100$ pixel ROI across the thirteen wavelength bands. Values are min-max normalized column wise (per wavelength band) for visualization.}
    \label{spectal_sig_classes}
\end{figure*}

\begin{figure*}[tb]
    \centering
    \includegraphics[width=\textwidth]{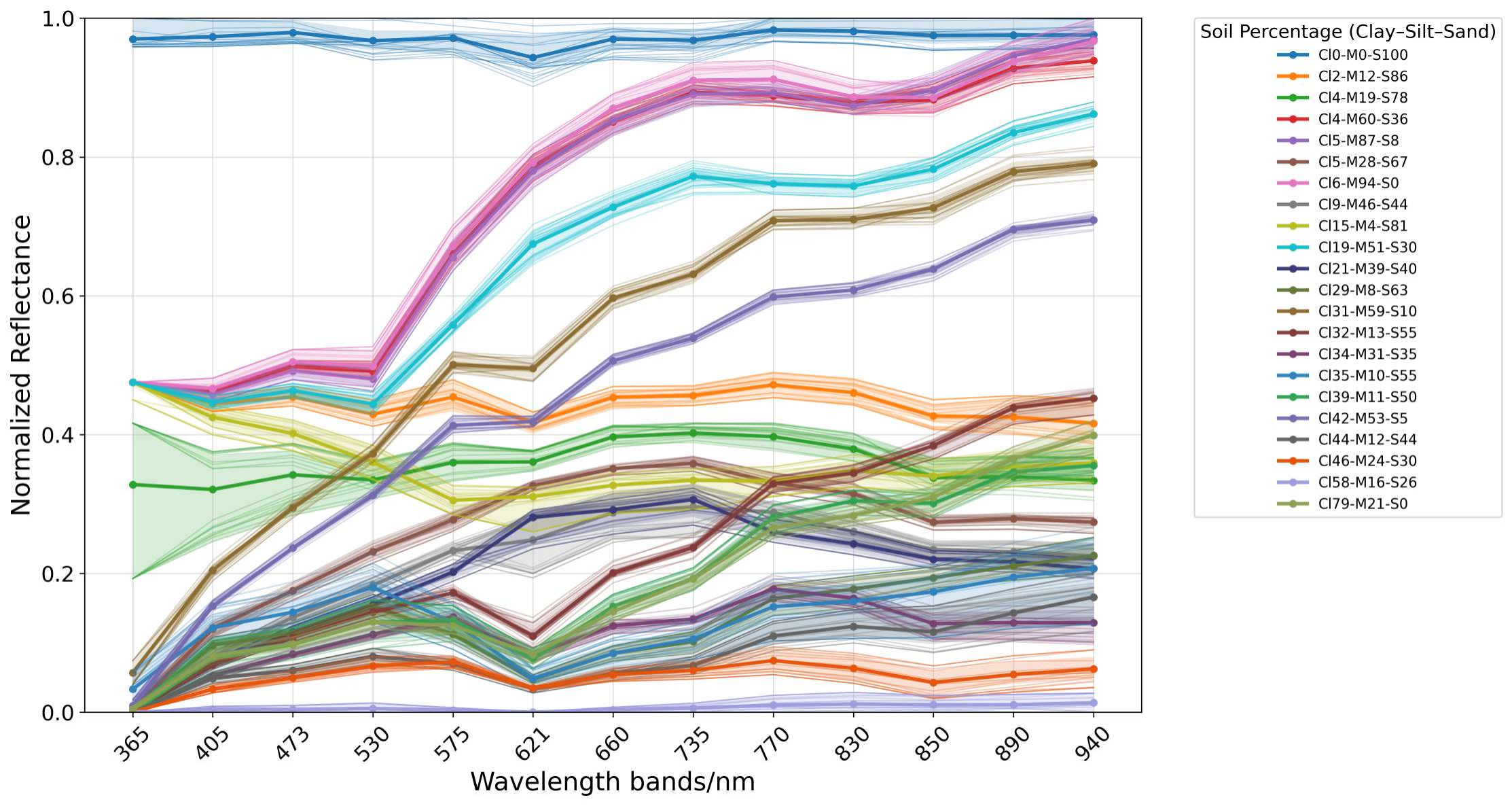}
    \caption{Spectral signatures grouped by composition levels (clay, silt, and sand) using the training/testing dataset. Each curve represents the mean intensity of the $100{\times}100$ pixel ROI across thirteen wavelength bands. Values are min-max normalized column wise (per wavelength band) for visualization. Abbreviations in the legend denote Clay (Cl), Silt (M), and Sand (S).}
    \label{spectal_sig_percentages}
\end{figure*}

To examine wavelength dependent reflectance variations and gain qualitative insight into spectral separability, spectral signatures were generated using the normalized training/testing dataset. For each wavelength band, average intensity values were computed and used to produce two types of spectral signatures.
The first spectral signature was obtained by plotting the average intensity values against the wavelength bands for all soil types, in order to visualize class wise trends and qualitative separability (Fig.~\ref{spectal_sig_classes}). The second spectral signature was obtained by plotting the average intensity values against the wavelength bands for all soil composition levels, to examine how changes in individual texture fractions influence the spectral response (Fig.~\ref{spectal_sig_percentages}).

Overall, these plots provide qualitative validation of the selected wavelength set and motivate the subsequent Linear Discriminant Analysis (LDA) based dimensionality reduction and machine learning models.

\subsection{Dimensionality Reduction}
\label{subs:dim_red}
Captured multispectral measurements can exhibit noticeable correlation across bands due to spectral overlap between adjacent wavelengths. To obtain a compact and more discriminative representation prior to learning, LDA was employed as a supervised dimensionality reduction step. The dataset prepared as described in Section~\ref{subs:feat_ext} and the LDA transform was fitted on the training subset only and then applied to the test and external validation sets using the same projection parameters to avoid information leakage.

LDA seeks a linear projection that maximizes class separability by maximizing the distance between-class scatters while minimizing the distance within-class scatters. Given $C$ classes, and feature vectors $\mathbf{x}\in\mathbb{R}^{d}$ ($d=13$) where each $\mathbf{x}$ represents a sample belonging to class $c$, the within-class scatter matrix ($\mathbf{S}_W$) and between-class scatter matrix ($\mathbf{S}_B$) are defined as,

\begin{equation}
\mathbf{S}_W=\sum_{c=1}^{C}\sum_{\mathbf{x}\in \mathcal{D}_c}(\mathbf{x}-\boldsymbol{\mu}_c)(\mathbf{x}-\boldsymbol{\mu}_c)^{\mathsf{T}},
\label{eq:sw}
\end{equation}
\begin{equation}
\mathbf{S}_B=\sum_{c=1}^{C} N_c(\boldsymbol{\mu}_c-\boldsymbol{\mu})(\boldsymbol{\mu}_c-\boldsymbol{\mu})^{\mathsf{T}},
\label{eq:sb}
\end{equation}

where $\mathcal{D}_c$ denotes the set of samples in class $c$, $N_c$ is the number of samples in class $c$, $\boldsymbol{\mu}_c$ is the class mean, and $\boldsymbol{\mu}$ is the global mean.

The discriminant directions are obtained by solving,

\begin{equation}
\mathbf{S}_B\mathbf{w}_k=\lambda_k \mathbf{S}_W\mathbf{w}_k,
\label{eq:lda_eig}
\end{equation}

where $\mathbf{w}_k$ is the eigenvector associated with the $k^{th}$ largest eigenvalue ($\lambda_k$). The eigenvectors are ordered corresponding to the eigenvalues in descending order ($\lambda_1\ge \lambda_2\ge \cdots$), and the projection matrix $\mathbf{W}=[\mathbf{w}_1,\ldots,\mathbf{w}_K]$ is formed by selecting the top $K$ directions, with $K\le \min(C-1,d)$. In this study, $K$ was selected to retain $99\%$ of the cumulative discriminative power (based on the ordered generalized eigenvalues), resulting in five components. As evaluation was done using five-fold cross-validation (Section\ref{sub:soil_tex_charac}),  the training portion of the dataset (four folds), $\mathbf{X}\in\mathbb{R}^{N\times 13}$ ($N = 35200$) was mapped into the LDA subspace using the learned projection matrix as,

\begin{equation}
\mathbf{Z}=\mathbf{X}\mathbf{W},
\label{eq:lda_proj_dataset}
\end{equation}

where $\mathbf{Z}\in\mathbb{R}^{N\times 5}$ ($N = 35200$) denotes the resulting low dimensional representation. The same transform with the fixed $\mathbf{W}$ learned from the training subset was then applied to the corresponding test potion of the each fold and external validation sets, and the resulting LDA features were used as inputs to the downstream models.

The supervisory classes used to learn the LDA projection depend on the analysis objective. For the soil type analysis, the USDA soil texture class labels were used to fit the LDA transform. For the composition analysis, samples were assigned to composition groups based on their clay, silt, and sand percentages (grouped composition levels), and these composition group labels were used as supervisory information to learn a separate LDA transform. The resulting LDA features were used as inputs to the downstream models for the following soil texture characterization strategies: 

\begin{enumerate}
    \item Direct classification of USDA soil texture classes (Section~\ref{subsub:strat1_direct}).
    \item Regression based estimation of clay, silt, and sand percentages (Section~\ref{subsub:strat2_regression}).
    \item Indirect USDA soil texture classification obtained by mapping the predicted composition from (2) using the USDA texture triangle decision rules (Section~\ref{subsub:strat3_indirect}).
\end{enumerate} 

\begin{figure}[tb]
    \centering
    \includegraphics[width=\linewidth]{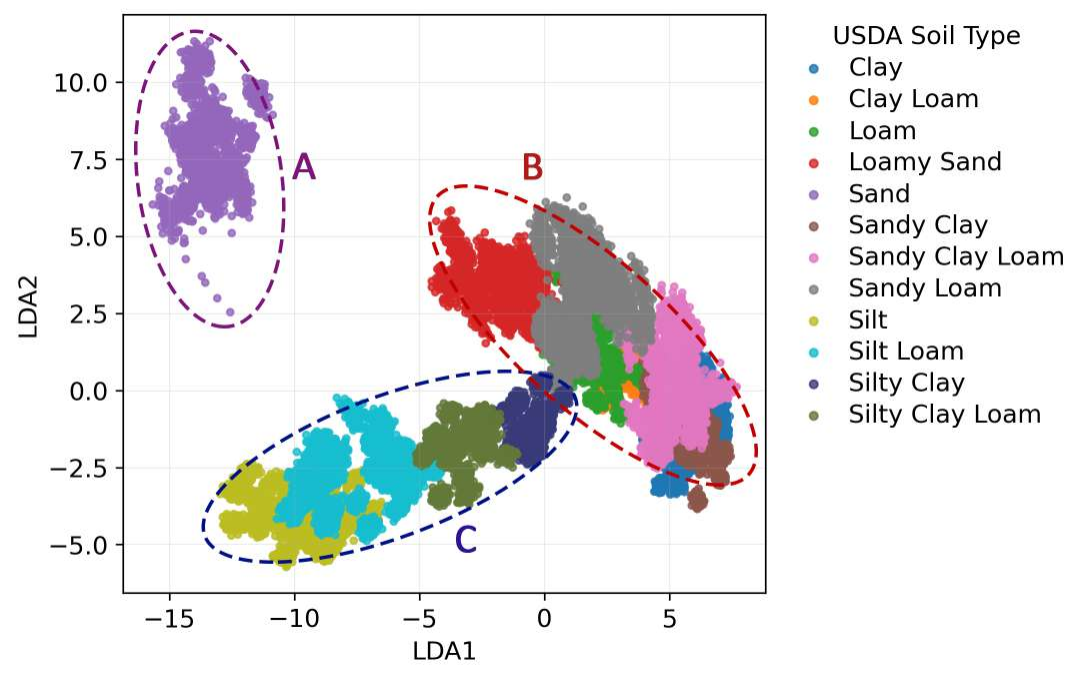}
    \caption{Two dimensional LDA visualization (LDA1 vs.\ LDA2) of the training/testing dataset, learned using USDA soil texture class labels. Dashed ellipses indicate three macro clusters: A (sand dominant textures), B (clay/loam dominant textures), and C (silt dominant textures).}
    \label{fig:lda_soiltype}
\end{figure}

\begin{figure}[tb]
    \centering
    \includegraphics[width=\linewidth]{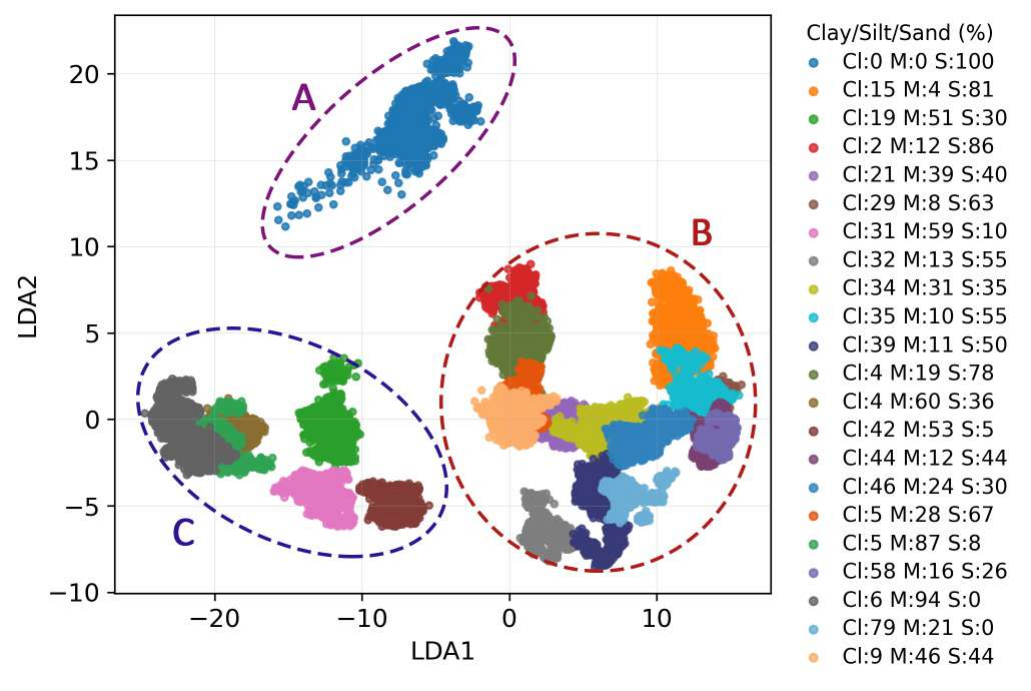}
    \caption{Two dimensional LDA visualization (LDA1 vs.\ LDA2) of the training/testing dataset, learned using composition group labels derived from clay (Cl), silt (M), and sand (S) percentages. Dashed ellipses indicate three macro clusters: A (sand dominant mixtures), B (clay/loam dominant mixtures), and C (silt dominant mixtures).}
    \label{fig:lda_comp}
\end{figure}

To qualitatively illustrate separability in the projected space, two dimensional scatter plots were generated using the first two discriminant components (LDA1 and LDA2). Fig.~\ref{fig:lda_soiltype} shows the projection learned using USDA soil texture class labels, whereas Fig.~\ref{fig:lda_comp} shows the projection learned using composition group labels derived from clay, silt, and sand percentages. In both plots, samples form compact, well separated clusters with only limited overlap among neighboring classes/composition groups, indicating that the MSI feature space is highly discriminative for texture assessment. Notably, both projections exhibit three dominant groupings that broadly align with texture triangle endmembers, a distinct sand rich cluster, a silt dominant cluster, and a broader mixed/clay-loam cluster. This trend is consistent with established soil reflectance behavior, where sand rich (often quartz dominated) soils typically show relatively high VNIR reflectance and low absorption, while silt and clay dominated soils display more absorption in the NIR to short-wave infrared (SWIR) ranges~\cite{DiRaimo2022,MadeiraNetto1996}.

\subsection{Soil Texture Characterization Strategies}
\label{sub:soil_tex_charac}
This section describes the learning setup and evaluation protocol used for each strategy. All models were trained and evaluated using the same LDA projected feature representation described in Section~\ref{subs:dim_red}. Model performance was assessed using five-fold cross-validation on the training/testing dataset. The complete dataset, comprising 44,000 datapoints, was partitioned into five mutually exclusive folds of approximately equal size. Evaluation was conducted over five runs, each using four folds (35,200 datapoints) for training and the remaining fold (8,800 datapoints) as a held-out test set. The reported performance corresponds to the mean and standard deviation across the five runs.

\subsubsection{Direct Soil Texture Class Classification}
\label{subsub:strat1_direct}
In the direct approach, the goal was to map multispectral measurements to USDA soil texture classes in a single step. The dataset was prepared as described in Section~\ref{subs:feat_ext}, and the corresponding USDA texture class labels were used as targets for supervised classification.

The feature vectors projected using LDA (Section~\ref{subs:dim_red}) were used as inputs to the downstream classifiers. During model training, class balancing was applied within each cross-validation fold. Synthetic Minority Over-sampling Technique (SMOTE)~\cite{smote} was applied only on the training portion of each fold to mitigate any class imbalance effects at the block level.

Several supervised classifiers were evaluated, including K-Nearest Neighbors (KNN), Random Forest (RF), Decision Trees (DT), CatBoost (CB), and XGBoost (XGB). For each fold, predictions were evaluated using overall accuracy, macro F1-score, and macro recall. Normalized confusion matrices were generated to visualize per-class behavior.

\subsubsection{Soil Composition Prediction via Regression Analysis}
\label{subsub:strat2_regression}
In this approach, the objective was to estimate soil composition from multispectral features by predicting the percentages of clay, silt, and sand. The same dataset as described in Section~\ref{subs:feat_ext} were used as inputs, and the regression targets were the laboratory derived clay, silt, and sand fractions.

The resulting LDA features were used to train and evaluate multiple regression models. KNN, RF, DT, XGB, and CB regressors were selected. Regression performance was quantified using the coefficient of determination ($R^2$) and Root Mean Squared Error (RMSE), computed separately for each composition component (clay, silt, and sand). Furthermore, the regressors were evaluated on the external validation dataset as described in Section~\ref{subs:feat_ext}, after applying the same LDA transformation (Section~\ref{subs:dim_red}) used for the training/testing data.

\subsubsection{Indirect Soil Texture Class Classification via Texture Triangle Mapping}
\label{subsub:strat3_indirect}
Here, indirect soil texture classification was performed by mapping the composition estimated through regression analysis (Section~\ref{subsub:strat2_regression}) to USDA soil texture classes. Specifically, each predicted (clay, silt, and sand) triplet was converted to a soil texture class using a rule based implementation of USDA texture triangle decision rules.

Indirect classification performance was evaluated by comparing the resulting USDA class predictions against the ground truth USDA labels, using accuracy, macro F1-score, and macro recall.

\section{Results}
\label{sec:results}

\subsection{Direct Soil Texture Class Classification}
\label{subs:res_direct_class}
Following the evaluation protocol described in Section~\ref{subsub:strat1_direct}, Table~\ref{tab:cls_test_direct} reports the five-fold cross-validation results of the direct soil texture classification models in terms of accuracy, macro F1-score, and macro recall (mean $\pm$ standard deviation).

\begin{table*}[tb]
\centering
\small
\caption{Direct soil-type classification performance on the test set}
\label{tab:cls_test_direct}
\begin{tabular}{lcccccc}
\toprule
\multirow{2}{*}{Model} &
\multicolumn{2}{c}{Accuracy} &
\multicolumn{2}{c}{F1-score} &
\multicolumn{2}{c}{Recall} \\
\cmidrule(lr){2-3}\cmidrule(lr){4-5}\cmidrule(lr){6-7}
 & Mean & Std. & Mean & Std. & Mean & Std. \\
\midrule
KNN & \textbf{0.9955} & \textbf{0.0007} & \textbf{0.9960} & \textbf{0.0007} & \textbf{0.9966} & \textbf{0.0005} \\
RF       & 0.9942 & 0.0010 & 0.9946 & 0.0009 & 0.9947 & 0.0008 \\
DT      & 0.9819 & 0.0012 & 0.9835 & 0.0014 & 0.9838 & 0.0013\\
CB            & 0.9899 & 0.0010 & 0.9904 & 0.0011 & 0.9918 & 0.0006 \\
XGB             & 0.9951 & 0.0012 & 0.9955 & 0.0012 & 0.9956 & 0.0010 \\
\bottomrule
\end{tabular}
\end{table*}

Among the evaluated methods, KNN achieved the best overall performance, with a mean accuracy of 0.9955, a mean macro F1-score of 0.9960, and a mean macro recall of 0.9966. RF and XGB delivered comparably strong results, reaching mean accuracies of 0.9942 and 0.9951, respectively, and with macro F1-scores of 0.9946 and 0.9955. CB achieved slightly lower yet competitive performance. In contrast, DT achieved the weakest performance and exhibited the highest fold to fold variation.

\begin{figure}[tb]
    \centering
    \includegraphics[width=\linewidth]{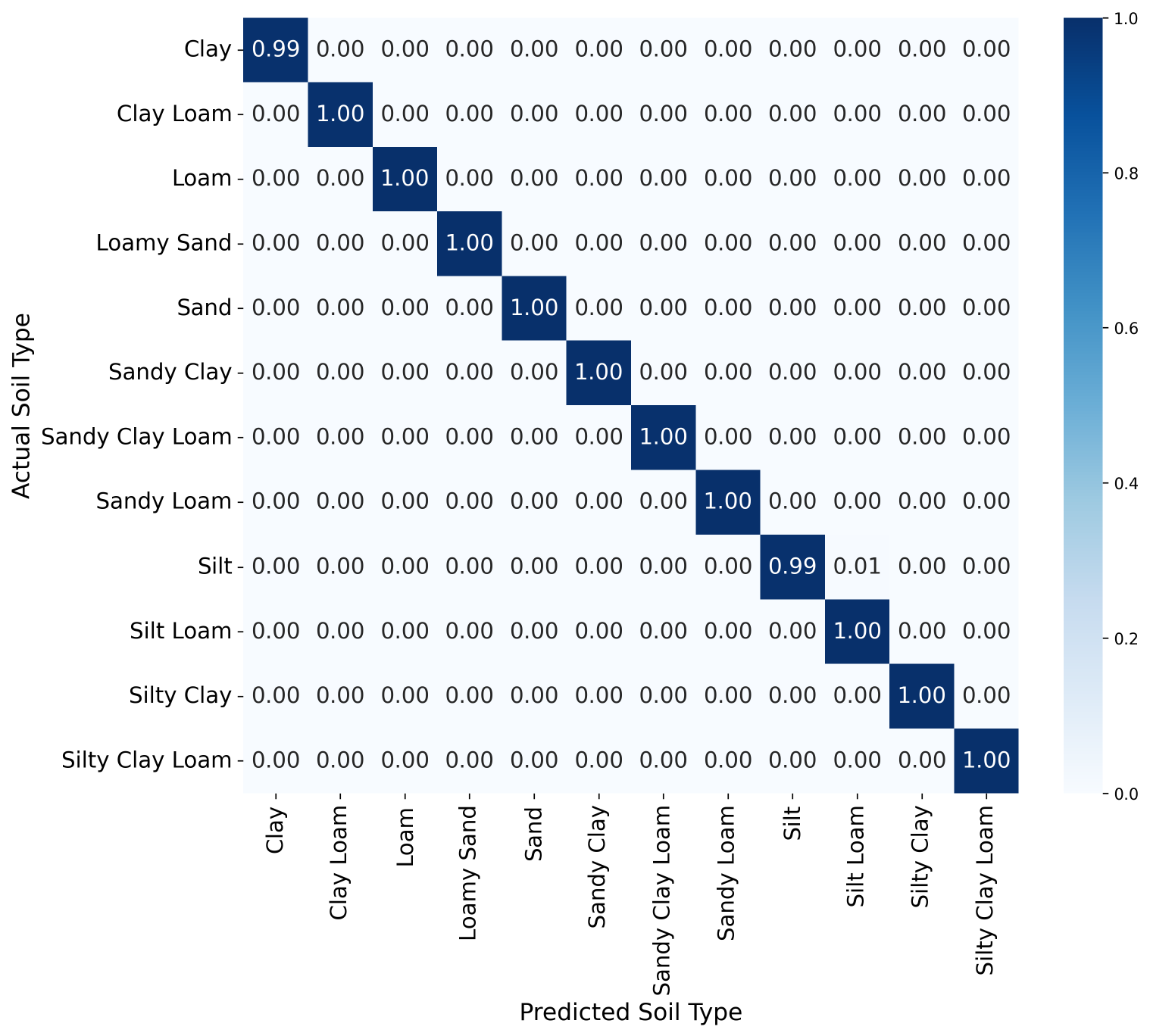}
    \caption{Normalized confusion matrix of the KNN classifier for direct soil classification.}
    \label{direct_cm}
\end{figure}

Overall, all models exhibit low standard deviations (in the order of $10^{-3}$), demonstrating strong robustness and repeatability across the cross-validation splits, with KNN showing the lowest variability. Furthermore, Fig.~\ref{direct_cm} presents the normalized confusion matrix for the best performing model (KNN) across the twelve soil classes.

\subsection{Soil Composition Prediction via Regression Analysis}
\label{subs:res_regr}
The methodology described in Section~\ref{subsub:strat2_regression} evaluates regression models for estimating clay, silt, and sand percentages from MSI features. Table~\ref{tab:reg_test} reports five-fold cross-validation results using $R^2$ and RMSE for each component (mean~$\pm$~standard deviation). Overall, all models achieved strong predictive accuracy, with $R^2$ values exceeding 0.986.

\begin{table*}[tb]
\centering
\caption{Regression results on held out testing data}
\label{tab:reg_test}
\small
\resizebox{\textwidth}{!}{%
\begin{tabular}{l 
                cc cc  
                cc cc  
                cc cc} 
\toprule
\multirow{3}{*}{Model} 
& \multicolumn{4}{c}{Clay} 
& \multicolumn{4}{c}{Silt} 
& \multicolumn{4}{c}{Sand} \\
\cmidrule(lr){2-5} \cmidrule(lr){6-9} \cmidrule(lr){10-13}
& \multicolumn{2}{c}{$R^2$} & \multicolumn{2}{c}{RMSE}
& \multicolumn{2}{c}{$R^2$} & \multicolumn{2}{c}{RMSE}
& \multicolumn{2}{c}{$R^2$} & \multicolumn{2}{c}{RMSE} \\
\cmidrule(lr){2-3} \cmidrule(lr){4-5}
\cmidrule(lr){6-7} \cmidrule(lr){8-9}
\cmidrule(lr){10-11} \cmidrule(lr){12-13}
& Mean & Std & Mean & Std 
& Mean & Std & Mean & Std 
& Mean & Std & Mean & Std \\
\midrule
RF   & 0.9956 & 0.0002 & 1.3525 & 0.0373 & 0.9957 & 0.0003 & 1.6936 & 0.0507 & 0.9926 & 0.0004 & 2.3963 & 0.0678 \\
KNN  & \textbf{0.9993} & \textbf{0.0001} & \textbf{0.5300} & \textbf{0.0279} & \textbf{0.9988} & \textbf{0.0001} & \textbf{0.9159} & \textbf{0.0469} & \textbf{0.9982} & \textbf{0.0002} & \textbf{1.1747} & \textbf{0.0569} \\
DT   & 0.9928 & 0.0008 & 1.7377 & 0.0980 & 0.9928 & 0.0004 & 2.1970 & 0.0617 & 0.9879 & 0.0004 & 3.0653 & 0.0560 \\
CB   & 0.9900 & 0.0003 & 2.0485 & 0.0357 & 0.9924 & 0.0004 & 2.2577 & 0.0517 & 0.9869 & 0.0006 & 3.1851 & 0.0715 \\
XGB  & 0.9946 & 0.0002 & 1.5020 & 0.0216 & 0.9952 & 0.0002 & 1.7979 & 0.0409 & 0.9918 & 0.0004 & 2.5273 & 0.0463 \\
\bottomrule
\end{tabular}
}
\end{table*}

\begin{figure*}[tb]
    \centering
    \includegraphics[width=1\linewidth]{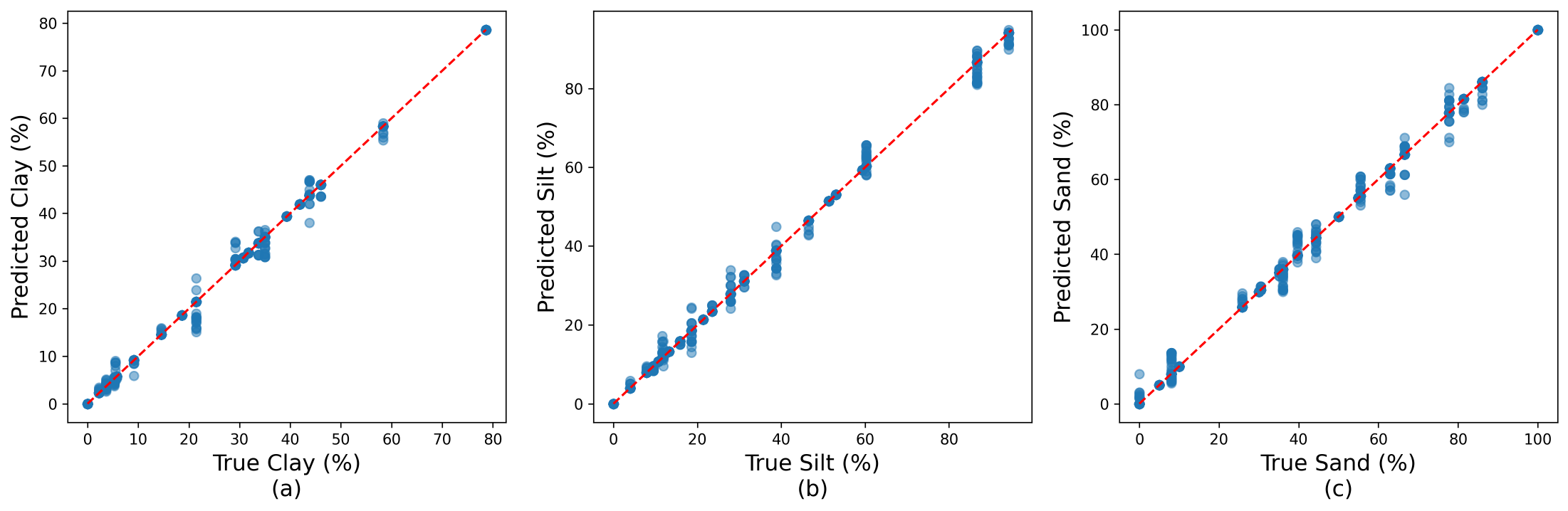}
    \caption{True vs predicted soil composition (\%) for the testing samples using the KNN regressor: (a) clay, (b) sand, and (c) silt.}
    \label{test_regression_plot}
\end{figure*}

\begin{figure*}[tb]
    \centering
    \includegraphics[width=1.0\linewidth]{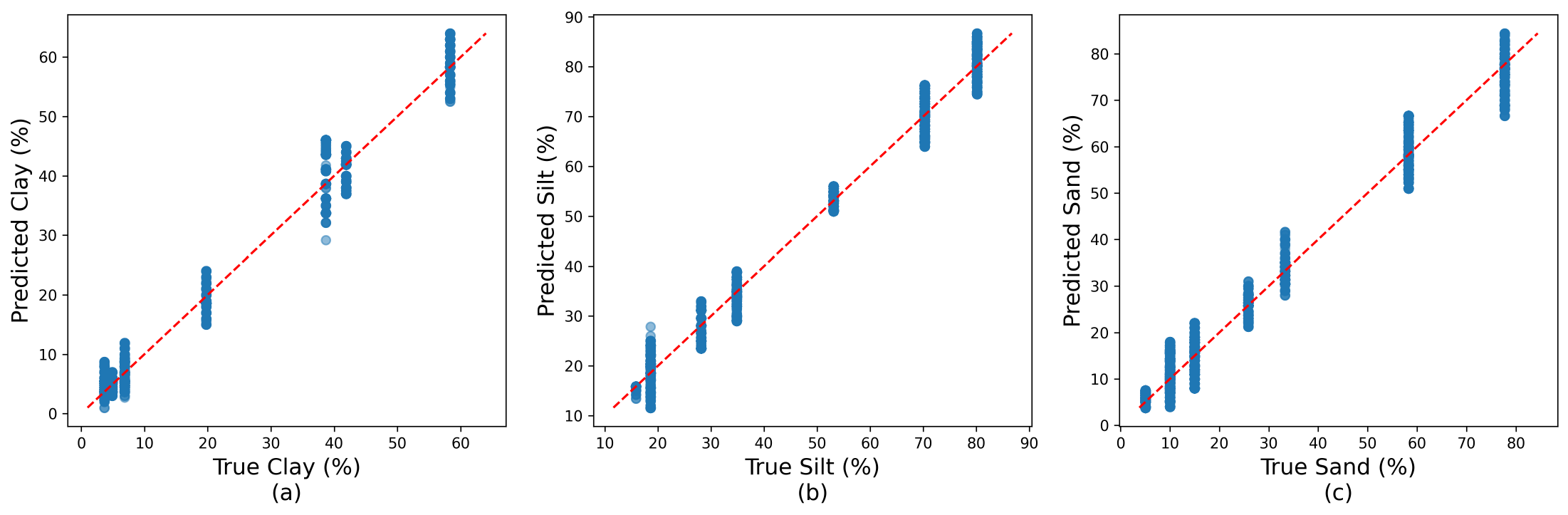}
    \caption {True vs predicted soil composition (\%) for the validation samples using the KNN regressor: (a) clay, (b) sand, and (c) silt.}
    \label{val_regression_plot}
\end{figure*}

Among the evaluated regressors, KNN achieved the best performance across all three components. It obtained a mean $R^2$ of 0.9993 with an RMSE of 0.5300 for clay,  a mean $R^2$ of 0.9988 with an RMSE of 0.9159 for silt, and a mean $R^2$ of 0.9982 with an RMSE of 1.1747 for sand. RF ranked second overall and produced consistent results, with mean $R^2$ values in the range of 0.9926-0.9957 across components. CB yielded the lowest performance, particularly for sand ($R^2$=0.9869, $RMSE$=3.1851). The standard deviations for both $R^2$ and RMSE are low across folds, indicating consistent performance across splits. The composition prediction outputs for the best performing regressor (KNN) on the testing data are visualized in Fig.~\ref{test_regression_plot}. Further, it was validated on the external validation dataset, and achieved $R^2$ of 0.9891 with an RMSE of 2.1238 for clay, a $R^2$ of 0.9801 with an RMSE of 3.2453 for sand, and an $R^2$ of 0.9891 with an RMSE of 2.5410 for silt. The prediction outputs for validation data are also graphically represented in Fig.~\ref{val_regression_plot}.

\subsection{Indirect Soil Texture Class Classification via Texture Triangle Mapping}
\label{subs:res_indirect}
Table~\ref{tab:cls_test_indirect} reports the five-fold cross-validation results of the indirect pipeline, where predicted clay, silt, and sand percentages are mapped to USDA texture classes using texture triangle decision rules. Performance is reported using accuracy, macro F1-score, and macro recall (mean $\pm$ standard deviation). The mean accuracy across models ranges from 0.9250 to 0.9698, with low standard deviations. Among the evaluated models, KNN achieved the best overall performance with a mean accuracy of 0.9698, a mean macro F1-score of 0.9682, and a mean macro recall of 0.9787. RF and DT produced comparable intermediate performance, whereas XGB yielded slightly lower results. CB showed the lowest performance, with a mean accuracy of 0.9250, a mean macro F1-score of 0.9226, and a mean macro recall of 0.9339.

\begin{table*}[tb]
\centering
\small
\caption{Indirect soil-type classification performance on the test set}
\label{tab:cls_test_indirect}
\begin{tabular}{lcccccc}
\toprule
\multirow{2}{*}{Model} &
\multicolumn{2}{c}{Accuracy} &
\multicolumn{2}{c}{F1-Score} &
\multicolumn{2}{c}{Recall} \\
\cmidrule(lr){2-3}\cmidrule(lr){4-5}\cmidrule(lr){6-7}
 & Mean & Std. & Mean & Std. & Mean & Std. \\
\midrule
KNN & \textbf{0.9698} & \textbf{0.0024} & \textbf{0.9682} & \textbf{0.0021} & \textbf{0.9787} & \textbf{0.0016} \\
RF       & 0.9650 & 0.0015 & 0.9652 & 0.0017 & 0.9724 & 0.0017 \\
DT       & 0.9560 & 0.0021 & 0.9550 & 0.0015 & 0.9659 & 0.0016\\
CB            & 0.9250 & 0.0032 & 0.9226 & 0.0029 & 0.9339 & 0.0025 \\
XGB             & 0.9455 & 0.0015 & 0.9427 & 0.0012 & 0.9568 & 0.0017 \\
\bottomrule
\end{tabular}
\end{table*}

\begin{figure}[tb]
    \centering
    \includegraphics[width=\linewidth]{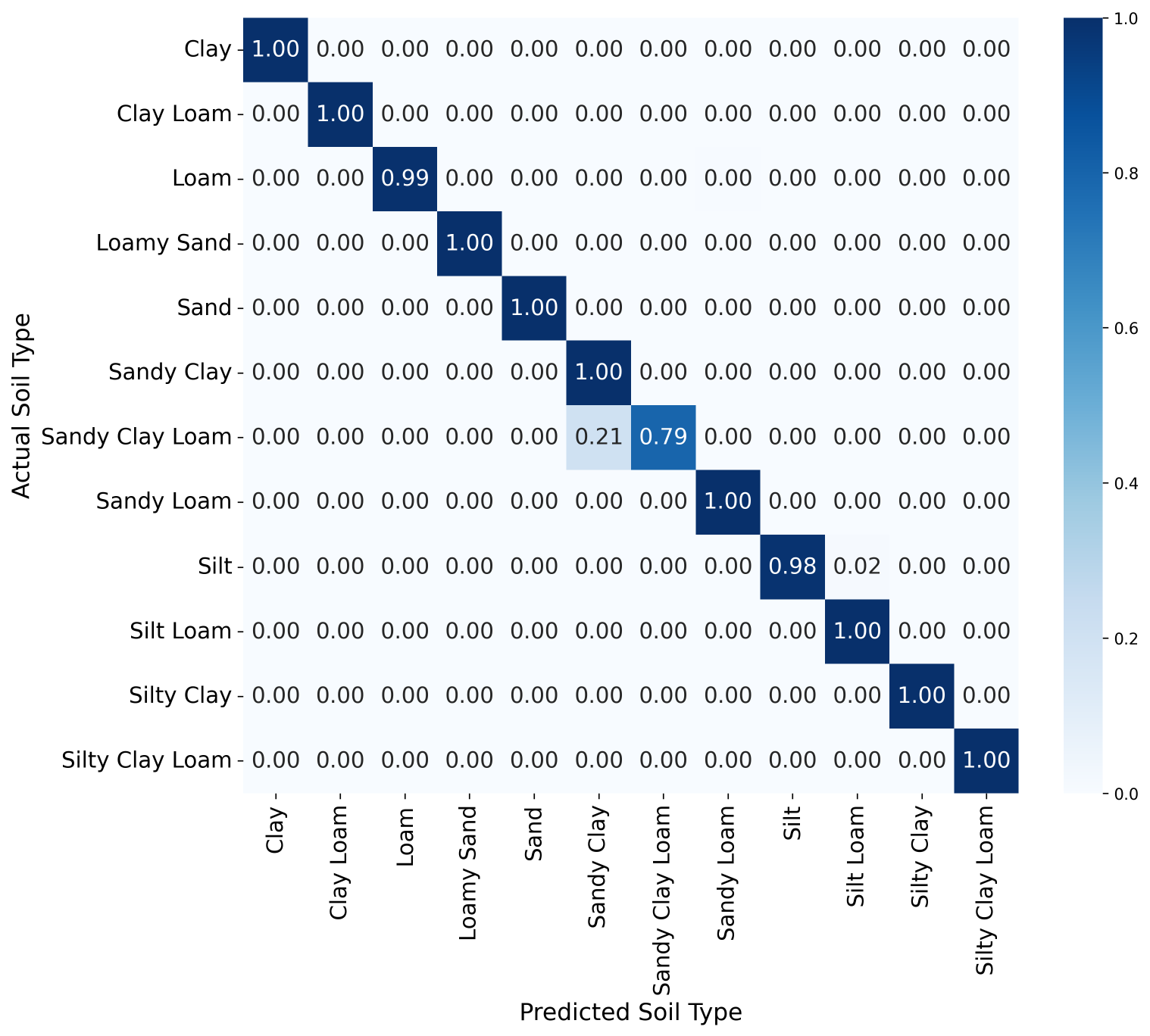}
    \caption{Normalized confusion matrix of the KNN model for indirect soil texture classification.}
    \label{indirect_cm}
\end{figure}

In addition, Fig.~\ref{indirect_cm} shows the normalized confusion matrix of the best performing indirect model (KNN). The most prominent misclassification occurs for Sandy Clay Loam, where 0.79 of the samples are correctly classified, and 0.21 are misclassified as Sandy Clay.

Overall, the best performing direct model (KNN, 0.9955 mean accuracy) exceeds the best performing indirect model (KNN, 0.9698 mean accuracy) by 0.0257 in absolute accuracy.

\section{Discussion}
\label{sec:discussion}

This section synthesizes the experimental findings to explain how and why different learning strategies behave under the same multispectral sensing setup. In particular, model performance and error patterns were interpreted for both the direct texture classification pipeline and the composition regression plus triangle mapping pipeline, and relate the observed differences to model stability and sensitivity to decision boundaries. The resulting trade-off between operational simplicity (one-step class prediction) and added interpretability (fraction estimates) is then highlighted, and implications for designing reliable, field deployable MSI and machine learning workflows are discussed.

\subsection{Performance Analysis of Direct Soil Texture Classification}
In the direct classification results (Table~\ref{tab:cls_test_direct}), all evaluated models achieve high performance, with the strongest methods approaching near perfect accuracy. This performance is expected as soils with different particle size distributions and constituent composition exhibit distinct, wavelength dependent reflectance responses in the utilized range. This is due to particle size controlled scattering and absorption by key components such as organic matter, iron oxides, and clay related constituents~\cite{Stenberg2010VisNIR,Sadeghi2018ParticleSize}. Consequently, the thirteen band multispectral feature space provides strong separability among the USDA textures. This enhanced separability explains the best overall performance of the KNN classifier, indicating that soil spectra form well structured clusters in the multispectral domain. Under this structure, samples from the same texture class lie close to one another, allowing a neighborhood based decision rule to classify effectively without requiring complex global partitioning.

RF and XGB also deliver comparably strong results. Their near matching performance indicates that ensemble learning captures nonlinear relationships and cross band interactions in multispectral reflectance, while providing robust generalization through bagging (RF) and boosting with regularization (XGB). In contrast, DT achieves the lowest performance and exhibits greater fold to fold variation, which is consistent with its high variance behavior. The small changes in the training data can lead to different split structures and less stable generalization across folds~\cite{Breiman2001RF,ChenGuestrin2016XGBoost}. CB remains competitive but trails the top performing methods, suggesting that, for this continuous spectral feature setting, its ordered boosting and categorical feature handling advantages are less pronounced than those of KNN and the strongest tree ensembles~\cite{Prokhorenkova2018CatBoost}. The close performance of macro F1-score and macro recall across the results further suggests that performance is not dominated by a subset of classes, but is broadly balanced across the class set.

The confusion matrix of the best performing KNN model provides additional insight into the remaining errors (Fig.~\ref{direct_cm}). The matrix is strongly diagonal, with most classes achieving perfect or near perfect classification. The most visible residual confusion occurs between Silt and Silt Loam, where a small fraction of Silt samples (approximately 0.01) is predicted as Silt Loam. These two classes are adjacent in the USDA texture triangle and can expect to exhibit similar spectral characteristics. This pattern suggests that the remaining errors are rare and are largely confined to neighboring texture categories rather than representing broad, systematic failure modes, further illustrating the robustness of the proposed direct classification pipeline.

\subsection{Comparative Analysis of Regression Models for Soil Composition Prediction}
In light of the regression analysis, the results (Table~\ref{tab:reg_test}) demonstrate that the proposed multispectral feature set supports accurate estimation of soil composition, as reflected by consistently high $R^2$ values and low RMSE across all evaluated models. These outcomes indicate that reflectance responses at the selected wavelengths are sensitive to texture related variation, enabling the regressors to learn stable mappings from spectral features to clay, silt, and sand percentages.

KNN achieves the lowest RMSE for all three components, suggesting that the relationship between multispectral signatures and soil composition is well captured by local structure in the feature space. Soils with similar compositions tend to exhibit similar spectral responses. Therefore, a neighborhood based regressor can interpolate composition by leveraging nearby samples rather than imposing a single global functional form. This local adaptability reduces prediction error and yields lower RMSE.

RF and XGB also produce strong performance, highlighting their ability to model nonlinear relationships and interactions among spectral bands. Their slightly higher RMSE compared with KNN may be attributed to the partition based nature of tree ensembles, where predictions are formed by averaging across terminal nodes. Such piecewise approximations can introduce mild bias, particularly in sparsely sampled regions of the composition space~\cite{Hastie2009ESL}. Although boosting and regularization can improve generalization, they may be less effective than local interpolation when the underlying mapping is smooth and neighborhood consistent~\cite{Stone1977Consistent}.

In contrast, CB underperforms relative to the other methods. One plausible explanation is that CB’s ordered boosting and categorical feature handling advantages are not fully utilized in a purely continuous spectral feature setting, making it less effective than KNN and other tree ensembles for learning fine grained spectral composition mappings~\cite{Prokhorenkova2018CatBoost}. DT also lags behind the top models. With multispectral inputs, discriminative information often arises from subtle band to band interactions and gradual spectral trends rather than sharp thresholds. As a result, a single tree can learn overly specific split rules that generalize less well across folds, leading to lower $R^2$ and higher RMSE~\cite{Breiman2001RF, Hastie2009ESL}. 

Across models, sand exhibits higher RMSE than clay and silt, indicating that the sand percentage is comparatively more challenging to estimate from the available spectral features. This is because the spectral signal in the utilized device range is driven mainly by active soil chromophores such as iron oxides, organic matter, and clay minerals. In contrast, the sand fraction is often dominated by quartz, which exhibits fewer pronounced absorption features within this spectral region~\cite{Pinheiro2017,Francos2021}. 

Overall, the consistently low RMSE achieved by KNN suggests that it provides the most reliable composition estimates for subsequent indirect soil texture classification using the USDA texture triangle, likely because the multispectral feature space yields well separated samples in composition space, making nearest-neighbor interpolation particularly effective. This is further confirmed by Fig.~\ref{test_regression_plot} and Fig.~\ref{val_regression_plot}, since for testing data and external validation data, the predictions of clay, silt, and sand lie closer to the ideal prediction line $y=x$. This indicates that the regression analysis generalized well to intermediate composition levels.

\subsection{Evaluation of Indirect Texture Classification via Composition Mapping}
Similar to the direct classification method, the indirect classification method also achieves high performance. Since KNN is capable of reliably separating the signatures associated with clay, silt, and sand, as indicated by its high $R^2$ and low RMSE values, its composition estimates can be mapped into the USDA texture triangle to extract soil texture classes with the highest accuracy, macro F1-score, and macro recall (Table~\ref{tab:cls_test_indirect}). By exploiting local similarity in the multispectral feature space, KNN produces composition estimates that more often remain within the correct USDA region, reducing boundary crossing errors during the texture triangle mapping step. The confusion matrix supports this interpretation (Fig.~\ref{indirect_cm}). The most notable misclassification occurs between Sandy Clay Loam and Sandy Clay, which occupy adjacent regions in the USDA texture triangle. For such closely related classes, relatively small deviations, particularly in the clay fraction, can shift a sample across a decision boundary and change the assigned texture class. Overall, the error patterns suggest that remaining mistakes are concentrated in samples located near USDA boundary lines.

\subsection{Performance Trade offs Between Direct and Indirect Classification Strategies}
Across experiments, the direct pipeline consistently outperforms the indirect pipeline. A key reason is that the indirect conversion amplifies small estimation errors due to the non-uniform geometric partitioning of the USDA texture triangle (Fig.~\ref{fig:texture_triangle}). The triangle is partitioned into regions of unequal size and shape with piecewise linear class boundaries. Consequently, the mapping becomes highly sensitive near boundaries, with small percentage errors in clay, silt, or sand causing a discrete class change even when the composition estimate is close. This is exemplified by the LDA projections in Fig.~\ref{fig:lda_soiltype} and Fig.~\ref{fig:lda_comp}, which show the clustering of USDA texture class labels and the clustering of clay, silt, and sand percentage compositions, respectively. If the estimated composition is slightly mismatched near a USDA triangle boundary, the sample can shift into a different composition cluster in the LDA space, and the subsequent USDA triangle mapping may assign a different soil texture class even when the prediction error is small.
Because this conversion is discontinuous, boundary adjacent samples are penalized more severely in the indirect approach, which helps explain why the direct strategy yields higher accuracy and improved stability.

Comparing direct and indirect soil texture classification pipelines is valuable because it clarifies where model complexity and interpretability should reside in practical deployments. The direct approach produces a USDA texture class in a single step, supporting fast, field ready decision support. The indirect approach yields interpretable composition estimates that can be integrated into agronomic workflows, such as fertilizer recommendations, irrigation scheduling, and soil amendment planning, and can be validated against laboratory measurements. The results, therefore, highlight a trade off between operational simplicity (direct) and interpretability with downstream utility (indirect), allowing practitioners to select an appropriate pipeline depending on whether the goal is rapid screening, monitoring, or detailed soil management.

This comparison is also scientifically meaningful because the USDA texture triangle is not an arbitrary partition. It is a physically meaningful geometric representation of the compositional constraint (clay + silt + sand = 100$\%$) and long established relationships between particle size fractions and soil behavior. In this sense, the triangle encodes soil science knowledge accumulated over time into a decision space, making it a versatile tool in diverse fields~\cite{USDA2017, Martin2018}. 

Our results indicate that, when provided with multispectral measurements, machine learning models can learn mappings that are consistent with this physical visual representation. Put differently, the pipeline recovers, from optical evidence, patterns that soil scientists historically distilled into the texture triangle. Crucially, the observed agreement between model outputs and USDA classes also depends on the quality of the captured signal. The results underscore the importance of the full pipeline (illumination design, band selection, acquisition consistency, preprocessing, and model training) in producing stable, discriminative reflectance features relevant to soil texture, enabling rapid and reliable deployment in the field.

\section{Conclusion}
\label{sec:conclusion}

This study proposes a novel, portable multispectral imaging (MSI) pipeline for soil texture characterization by the United States Department of Agriculture (USDA). Soil images were acquired using an in-house, low cost, and rapidly deployable MSI device equipped with thirteen narrowband spectral channels spanning 365-940 nm, each implemented using an array of LEDs. Three complementary strategies were evaluated: (i) direct prediction of USDA soil texture classes from multispectral features, (ii) soil composition estimation via regression analysis, and (iii) indirect texture classification by mapping the estimated soil composition to USDA texture classes using texture triangle decision rules. All three strategies were assessed under five-fold cross-validation. 

For the direct approach, several classifiers were tested, and K-Nearest Neighbors (KNN) achieved the best performance with a mean accuracy of 99.55\%, confirming the effectiveness of the proposed method for direct texture classification. The regression results further show that the selected spectral bands contain sufficient information to accurately estimate soil composition of clay, silt, and sand. For composition prediction, multiple regressors were evaluated, and KNN again performed best [Clay: $R^2{=}$0.9993, RMSE$\,{=}$0.5300; Silt: $R^2{=}$0.9988, RMSE$\,{=}$0.9159; Sand: $R^2{=}$0.9982, RMSE$\,{=}$1.1747]. The regression models were further validated on an external dataset [Clay: $R^2{=}$0.9891, RMSE$\,{=}$2.1238; Silt: $R^2{=}$0.9891, RMSE$\,{=}$2.5410; Sand: $R^2{=}$0.9801, RMSE$\,{=}$3.2453]. These results indicate strong generalization performance, particularly for intermediate composition levels. In turn, these composition estimates enable indirect USDA texture classification via the texture triangle. For the indirect strategy, KNN achieved the highest accuracy of 96.98\% which is comparable to the direct method, albeit with a slight drop that could be attributed to the two step process and the linear boundaries of the USDA soil texture triangle.

Overall, the findings demonstrate that the proposed MSI feature space enables both accurate texture class prediction and independent soil composition estimation, within a single sensing framework. The portable multispectral platform exhibits robust performance across direct classification, composition prediction independent of classification, and indirect classification, while serving as a viable, highly scalable tool for rapid, low cost field screening and mass deployment. This capability supports consistent measurements across sites and offers clear potential for integration into routine agroecosystems, geotechnical practices, hydrological analyses, and environmental management advisory services. Collectively, the proposed imaging system and end to end workflow provide a practical bridge between classical soil texture knowledge and scalable, data-driven soil assessment.

\section*{Acknowledgment}

The authors gratefully acknowledge the staff of the Geotechnical Laboratory, Department of Civil Engineering, Faculty of Engineering, University of Peradeniya, for their support and assistance throughout this study.

\bibliographystyle{ieeetr}
\bibliography{paper-refs}

\newpage

\begin{IEEEbiography}[{\includegraphics[width=1in,height=1.25in,clip,keepaspectratio]{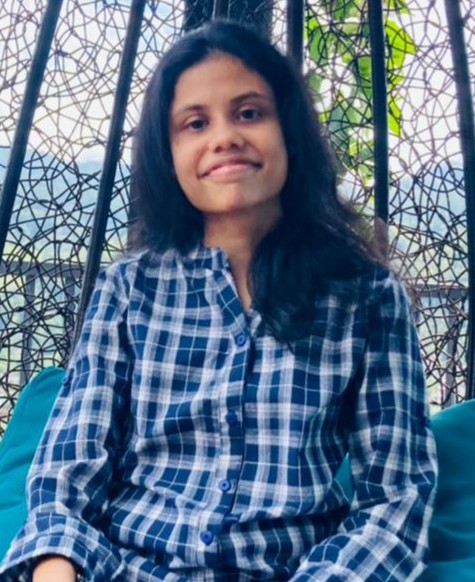}}]{G.A.S.L. Ranasinghe} received the B.Sc. (Eng.) degree (First Class Honors) in electrical and electronic engineering from the University of Peradeniya, Peradeniya, Sri Lanka. She is currently a volunteer researcher with the Multidisciplinary AI Research Centre (MARC), University of Peradeniya. She has published papers in multiple IEEE conferences. Her research interests include signal processing, image processing, deep learning, generative artificial intelligence, and applied mathematics.
\end{IEEEbiography}

\begin{IEEEbiography}[{\includegraphics[width=1in,height=1.25in,clip,keepaspectratio]{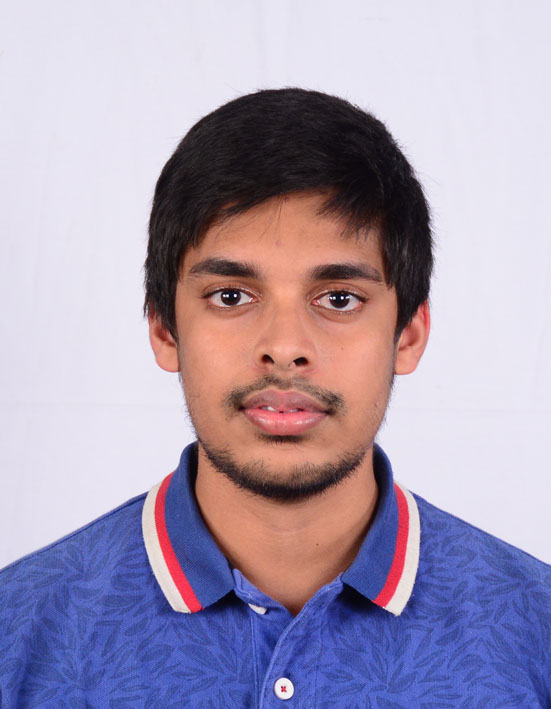}}]{J.A.S.T. Jayakody} received the B.Sc. (Eng.) degree (First Class Honors) in electrical and electronic engineering from the University of Peradeniya, Peradeniya, Sri Lanka. He is currently a Research Assistant with the Multidisciplinary AI Research Centre (MARC), University of Peradeniya. He has published papers in multiple IEEE conferences. His research interests include image processing, machine learning, signal processing, and computer vision for agricultural and environmental sensing applications.
\end{IEEEbiography}

\begin{IEEEbiography}[{\includegraphics[width=1in,height=1.25in,clip,keepaspectratio]{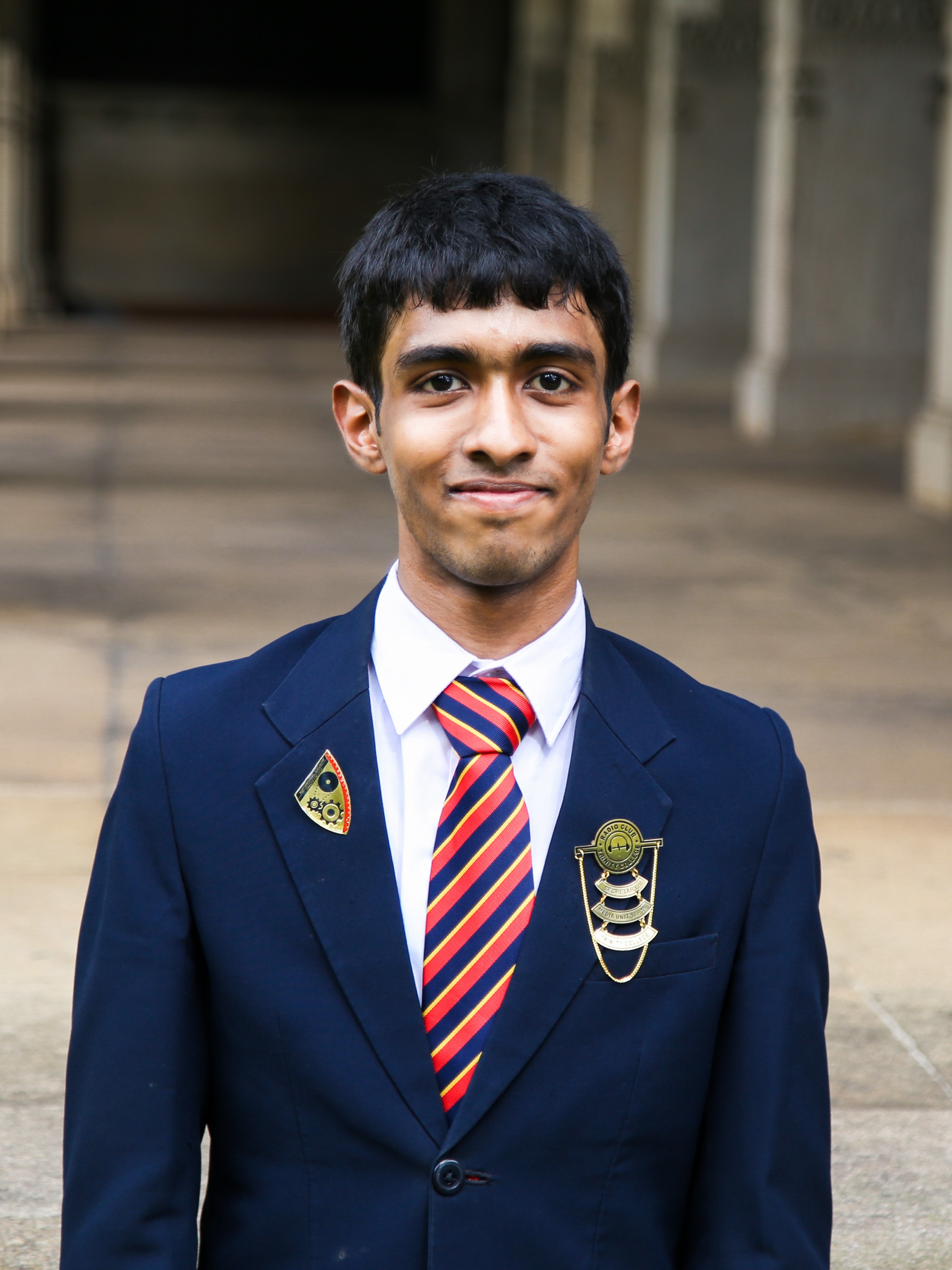}}]{M.C.L. De Silva} received the B.Sc. (Eng.) degree (First Class Honors) in electrical and electronic engineering from the University of Peradeniya, Sri Lanka, in 2025. He is currently a Volunteer Researcher with the Multidisciplinary AI Research Center (MARC), University of Peradeniya, contributing to projects in reflectance multispectral imaging for agricultural and environmental monitoring and data-driven agent-based epidemic simulation. His research interests include statistical and spectral signal processing, computational imaging, implicit neural representations, and applied statistics.
\end{IEEEbiography}

\begin{IEEEbiography}[{\includegraphics[width=1in,height=1.25in,clip,keepaspectratio]{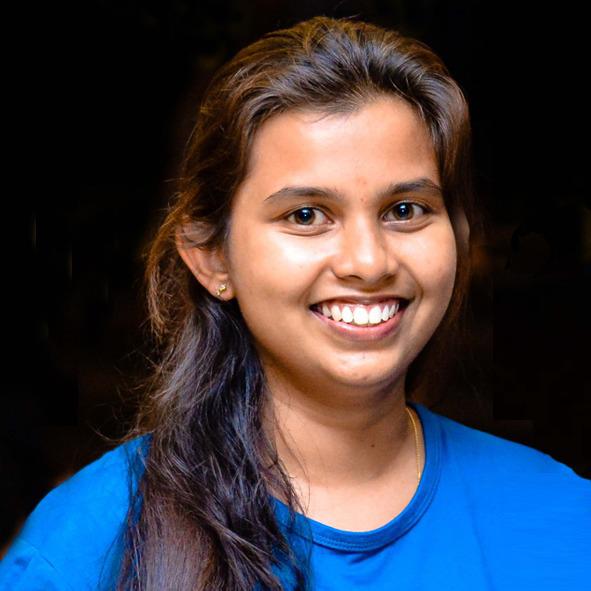}}]{G. Thilakarathne} received the B.Sc. (Special) degree (First Class Honors) in Applied Electronics from Wayamba University of Sri Lanka, Sri Lanka, in 2022. She is currently pursuing the Ph.D. degree with the Faculty of Engineering, University of Peradeniya, Sri Lanka. Her research focuses on multispectral imaging for condition evaluation in organic agriculture supply chains. Her research interests include image processing, machine learning, and electronic system development for image-based agricultural quality assessment and food safety applications.
\end{IEEEbiography}

\begin{IEEEbiography}[{\includegraphics[width=1in,height=1.25in,clip,keepaspectratio]{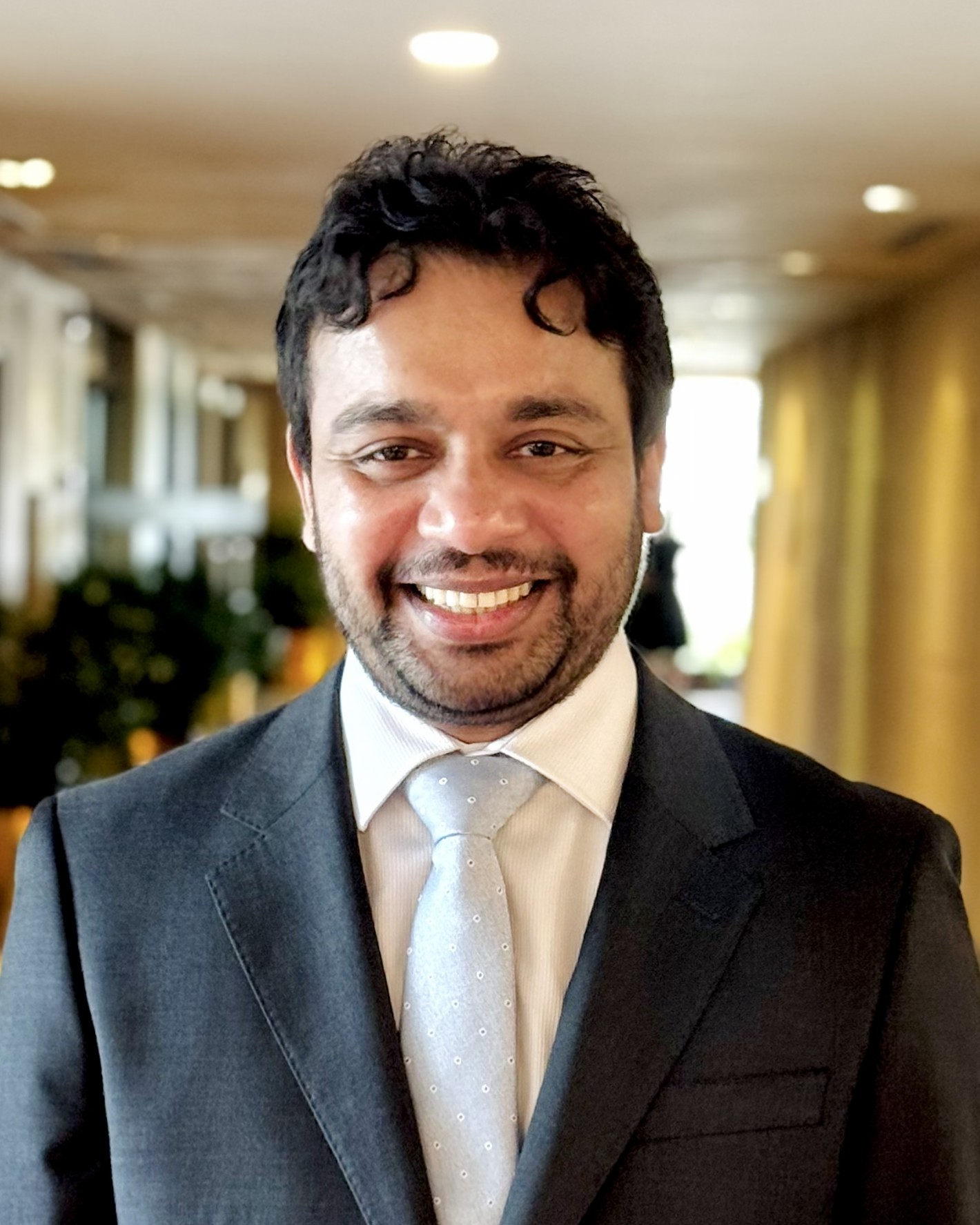}}]{G.M.R.I. Godaliyadda} (Senior Member, IEEE) received the B.Sc. (Eng.) degree (First Class Honors) in Electrical and Electronic Engineering from the University of Peradeniya, Sri Lanka, in 2005, and the Ph.D. degree in Electrical and Computer Engineering from the National University of Singapore, Singapore, in 2011.

He is currently a Professor with the Department of Electrical and Electronic Engineering, Faculty of Engineering, University of Peradeniya. His research interests include signal/image processing, machine learning, biomedical signal processing, wearable and non-intrusive sensing, and healthcare decision support with applications in multispectral/hyperspectral imaging, remote sensing and geoscience analytics, and smart grids.

Dr. Godaliyadda has received multiple Sri Lanka President’s Awards for Scientific Research, best paper awards, and competitive grants from NSF Sri Lanka and IDRC Canada. He is also an awardee of the NVIDIA Academic Grant Program supporting the HeritageAI initiative (8K A100 GPU-hours on Brev).
\end{IEEEbiography}

\begin{IEEEbiography}[{\includegraphics[width=1in,height=1.25in,clip,keepaspectratio]{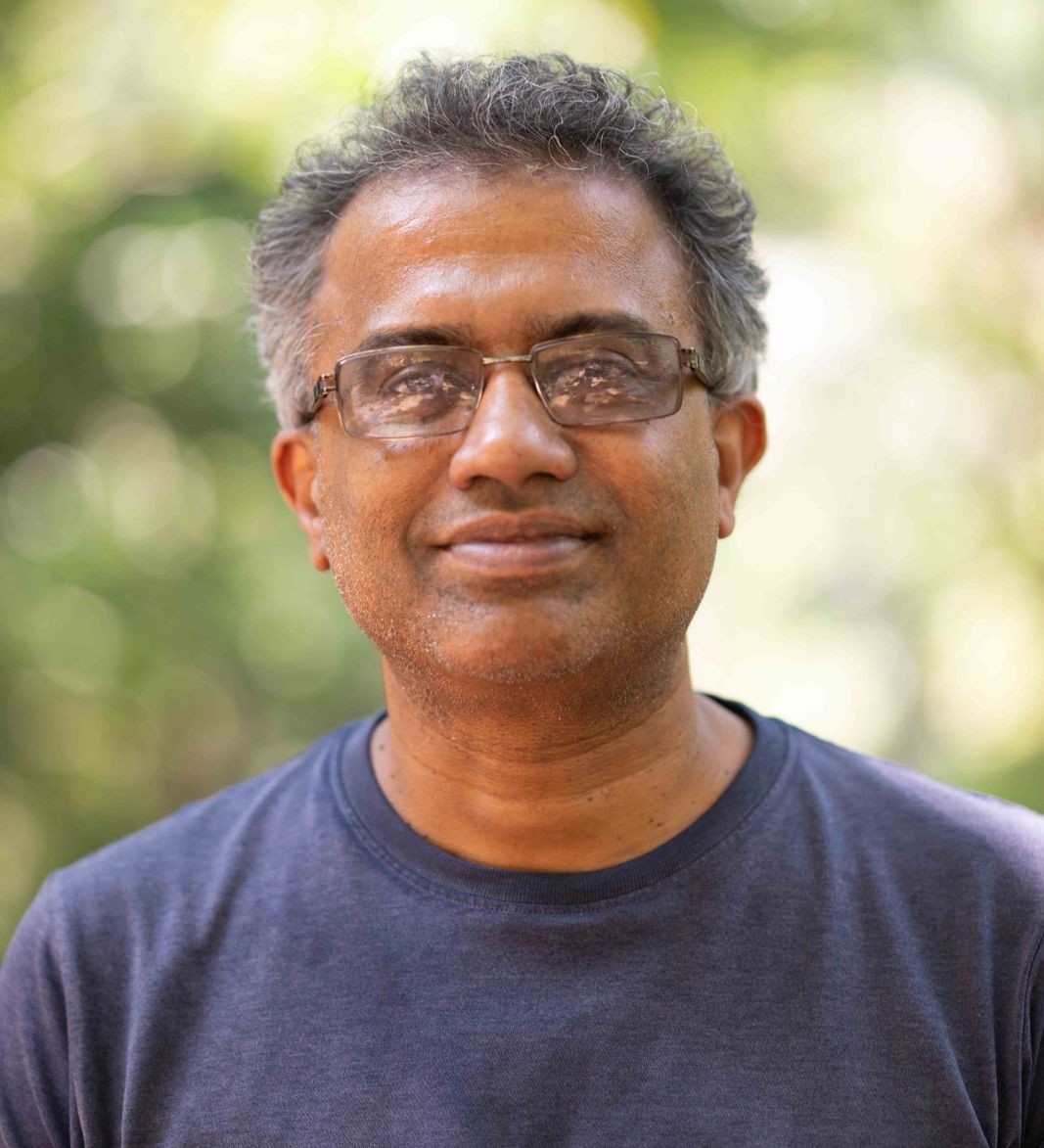}}]{H.M.V.R. Herath} (Senior Member, IEEE) received the B.Sc. Eng. degree in Electrical and Electronic Engineering with First-Class Honours from the University of Peradeniya, Sri Lanka, in 1998. He earned his M.Sc. degree in Electrical and Computer Engineering with academic merit from the University of Miami, USA, in 2002, and his Ph.D. in Electrical Engineering from the University of Paderborn, Germany, in 2009.

Currently, he is a Professor and the Head of the Department of Electrical and Electronic Engineering at the University of Peradeniya. His research interests include remote sensing, multispectral imaging, AI and machine learning, and light communication.

Prof. Herath is a Member of the Institution of Engineers, Sri Lanka (IESL), and a Senior Member of Optica. He served as the General Chair of the 2013 IEEE International Conference on Industrial and Information Systems (ICIIS) held in Kandy, Sri Lanka. He is also a recipient of best paper awards at the ICTer 2017 and MERCon 2023 conferences. Furthermore, he received the Sri Lanka President’s Award for Scientific Research in 2013 and 2020.
\end{IEEEbiography}

\begin{IEEEbiography}[{\includegraphics[width=1in,height=1.25in,clip,keepaspectratio]{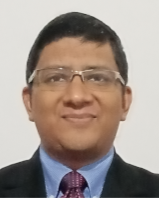}}]{M.P.B. Ekanayake} (Senior Member, IEEE) received the B.Sc. (Eng.) degree (First Class Honors) in Electrical and Electronic Engineering from the University of Peradeniya, Sri Lanka, in 2006, and the Ph.D. degree in Applied Mathematics from Texas Tech University, Lubbock, TX, USA, in 2011.

Currently, he is a Professor with the Department of Electrical and Electronic Engineering, University of Peradeniya. His research interests include biomedical signal processing, wearable and non-intrusive sensing, and data-driven modeling for maternal and fetal health, as well as AI and machine learning, computer vision, and multispectral/hyperspectral imaging, and data-driven methods for smart grids and complex systems.

Prof. Ekanayake has received multiple Sri Lanka President’s Awards for Scientific Publications, best paper awards at international conferences, and competitive research grants from the National Science Foundation (NSF), Sri Lanka, and other funding agencies. He is also a Co-Investigator of a multidisciplinary research project on preterm birth detection using electrohysterogram and tocogram signals.
\end{IEEEbiography}

\begin{IEEEbiography}[{\includegraphics[width=1in,height=1.25in,clip,keepaspectratio]{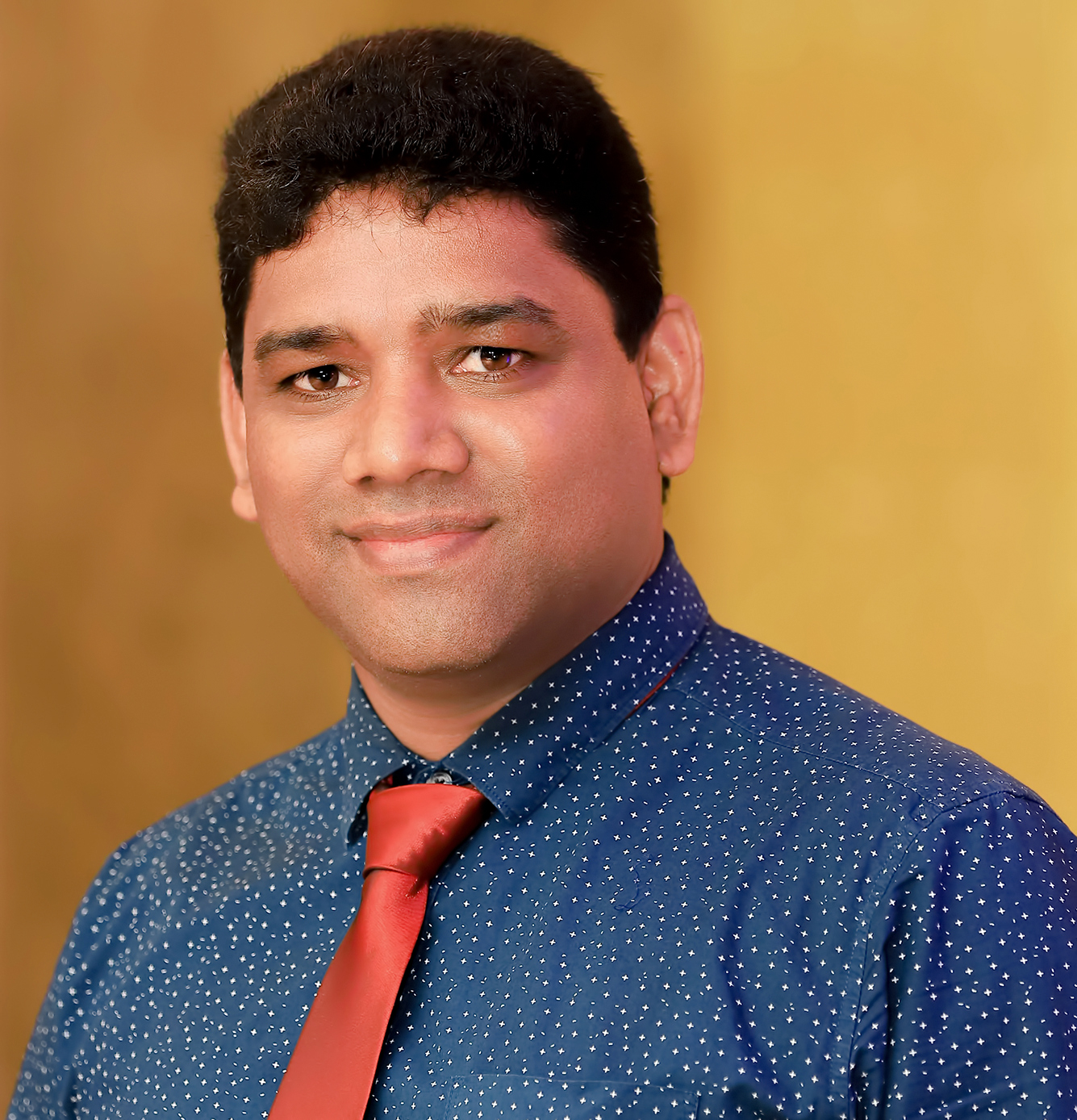}}]{S.K. Navarathnarajah} received the B.Sc. Eng. degree (First Class Honors) in Civil Engineering from the University of Peradeniya, Sri Lanka, in 2001, the M.Sc. degree in Civil Engineering (Geotechnical Engineering) from the University of Oklahoma, USA, in 2006, and the Ph.D. degree in Civil Engineering (Geotechnical Engineering) with Examiners’ Commendation for Outstanding Thesis from the University of Wollongong, Australia, in 2017. He is currently a Senior Lecturer with the Department of Civil Engineering, Faculty of Engineering, University of Peradeniya, Sri Lanka, where he also serves as the Postgraduate Coordinator for Geotechnical Engineering. Prior to academia, he worked as a Project and Staff Engineer with Group Delta Consultants, Inc., California, USA, and is a registered Professional Engineer (P.E., Civil) in the State of California. He has over 20 years of combined academic and industry experience. He has published widely in leading international journals and conferences. His research interests include railway geotechnics, ballasted rail track behavior under cyclic loading, ground improvement using recycled materials, pavement engineering, and advanced numerical modeling. He received the ICE-UK Best Journal Paper Award for Overseas Authors (Sir Mokshagundam Visvesvaraya Award) in 2023 and the President’s Award for Scientific Research, Sri Lanka, in 2018.
\end{IEEEbiography}

\EOD

\end{document}